\documentclass[10pt,twocolumn,letterpaper]{article}

\usepackage[pagenumbers]{cvpr} % To force page numbers, e.g. for an arXiv version

\usepackage{graphicx}
\usepackage{amsmath}
\usepackage{amssymb}
\usepackage{booktabs}

\usepackage{nicefrac}       %
\usepackage{microtype}      %

\usepackage{blindtext}
\usepackage{float}
\usepackage{enumitem}
\usepackage[table]{xcolor}
\usepackage{tabulary,multirow,overpic}
\usepackage{verbatim}
\usepackage{color}
\usepackage{multicol}
\usepackage{dblfloatfix}
\usepackage{pifont}%
\usepackage[accsupp]{axessibility}  %
\usepackage{makecell}

\usepackage{multicol}

\usepackage{fontawesome5}
\usepackage{textpos}

\makeatletter
\def\@fnsymbol#1{\ensuremath{\ifcase#1\or \dagger\or \ddagger\or
   \mathsection\or \mathparagraph\or \|\or **\or \dagger\dagger
   \or \ddagger\ddagger \else\@ctrerr\fi}}
\makeatother

\newlength\savewidth\newcommand\shline{\noalign{\global\savewidth\arrayrulewidth
\global\arrayrulewidth 1pt}\hline\noalign{\global\arrayrulewidth\savewidth}}
\newcommand{\tablestyle}[2]{\setlength{\tabcolsep}{#1}\renewcommand{\arraystretch}{#2}\centering\footnotesize}

\newcolumntype{x}[1]{>{\centering\arraybackslash}p{#1pt}}
\newcolumntype{y}[1]{>{\raggedright\arraybackslash}p{#1pt}}
\newcolumntype{z}[1]{>{\raggedleft\arraybackslash}p{#1pt}}

\definecolor{baselinecolor}{gray}{.92}

\definecolor{demphcolor}{gray}{.2}

\definecolor{demphcolorinline}{gray}{.4}
\newcommand{\demphinline}[1]{\textcolor{demphcolorinline}{#1}}

\definecolor{demphcolor1}{gray}{.6}
\newcommand{\demphs}[1]{\textcolor{demphcolor1}{#1}}

\definecolor{evaunit01purple}{RGB}{168,119,200}
\newcommand{\evapurple}[1]{\textcolor{evaunit01purple}{#1}}

\definecolor{evaunit01green}{RGB}{82,208,83}
\newcommand{\evagreen}[1]{\textcolor{evaunit01green}{#1}}

\definecolor{evaunit02red}{RGB}{211,41,15}

\definecolor{eva02yellow}{RGB}{230,119,11}

\newcommand{\ph}[1]{\textcolor{white}{#1}}

\definecolor{citecolor}{RGB}{34,139,34}
\definecolor{citecolor2}{HTML}{0071bc}
\definecolor{Graylight}{gray}{0.9}
\definecolor{lightred}{RGB}{241,140,142}
\usepackage[pagebackref=true,breaklinks=true,colorlinks,
citecolor=evaunit01green,urlcolor=evaunit01purple,bookmarks=false]{hyperref}

\definecolor{clipbaselinecolor}{gray}{.9}

\definecolor{defaultcolor}{HTML}{E8E2F7}
\newcommand{\evadefault}[1]{\cellcolor{defaultcolor}{#1}}

\renewcommand{\paragraph}[1]{\vspace{1.25mm}\noindent\textbf{#1}}

\newcommand{\app}{\raise.17ex\hbox{$\scriptstyle\sim$}}

\def\x{$\times$}
\newcommand{\figref}[1]{Fig.~\ref{#1}}
\newcommand{\tblref}[1]{Table~\ref{#1}}
\newcommand{\sref}[1]{\S\ref{#1}}

\newcommand{\boxAP}{AP$^\text{box}$\xspace}
\newcommand{\maskAP}{AP$^\text{mask}$\xspace}
\newcommand{\maskAPrare}{AP$^\text{mask}_\text{rare}$\xspace}
\newcommand{\maskAPcomm}{AP$^\text{mask}_\text{comm}$\xspace}
\newcommand{\maskAPfreq}{AP$^\text{mask}_\text{freq}$\xspace}
\newcommand{\boundaryAPfix}{AP$^\text{boundary}$\xspace}
\newcommand{\val}{$\mathtt{val}$\xspace}
\newcommand{\test}{$\mathtt{test}$-$\mathtt{dev}$\xspace}

\newcommand{\mIoUss}{mIoU$^\text{ss}$\xspace}
\newcommand{\mIoUms}{mIoU$^\text{ms}$\xspace}

\newcommand{\dt}[1]{\fontsize{6pt}{0.1em}\selectfont (#1)}
\newcommand{\dtplus}[1]{\fontsize{6pt}{0.1em}\selectfont (\textbf{\evagreen{#1}})}

\newcommand{\eva}{{\textbf{\evapurple{EVA}}}\xspace}
\newcommand{\EVA}{{\textbf{\evapurple{EVA}}}\xspace}

\newcommand{\suptext}[1]{$^{\text{#1}}$}

\newcommand{\cmark}{\ding{51}}%
\newcommand{\xmark}{\ding{55}}%
\newcommand{\authorskip}{\hspace{5mm}}

\usepackage[capitalize]{cleveref}
\crefname{section}{Sec.}{Secs.}
\Crefname{section}{Section}{Sections}
\Crefname{table}{Table}{Tables}
\crefname{table}{Tab.}{Tabs.}

\def\cvprPaperID{1997} % *** Enter the CVPR Paper ID here
\def\confName{CVPR}
\def\confYear{2023}

\begin{document}

\title{\eva: Exploring the Limits of Masked Visual Representation Learning at Scale}

\author{Yuxin Fang\textsuperscript{2,1}\thanks{Interns at Beijing Academy of Artificial Intelligence.} \authorskip Wen Wang\textsuperscript{3,1}$^\dag$ \authorskip Binhui Xie\textsuperscript{4,1}$^\dag$ \authorskip Quan Sun\textsuperscript{1} \authorskip Ledell Wu\textsuperscript{1} \\[0.5mm]
Xinggang Wang\textsuperscript{2} \authorskip Tiejun Huang\textsuperscript{1} \authorskip Xinlong Wang\textsuperscript{1} \authorskip Yue Cao\textsuperscript{1} \\[2mm]
{
\fontsize{10.4pt}{9.84pt}\selectfont
\textsuperscript{1}Beijing Academy of Artificial Intelligence \hspace{5.5mm} \textsuperscript{2}Huazhong University of Science and Technology 
}\\
{
\fontsize{10.4pt}{9.84pt}\selectfont
\ph{++...}\textsuperscript{3}Zhejiang University \hspace{5.7mm} \textsuperscript{4}Beijing Institute of Technology
}\\[1.5mm]
{
\fontsize{9.4pt}{9.84pt}\selectfont
Code \& Models: \url{https://github.com/baaivision/EVA}
}
}

\maketitle

\begin{abstract}
   We launch \eva, a vision-centric foundation model to \textbf{\evapurple{E}}xplore the limits of \textbf{\evapurple{V}}isual representation at sc\textbf{\evapurple{A}}le using only publicly accessible data. \EVA is a vanilla ViT pre-trained to reconstruct the masked out image-text aligned vision features conditioned on visible image patches. Via this pretext task, we can efficiently scale up \eva to one billion parameters, and sets new records on a broad range of representative vision downstream tasks, such as image recognition, video action recognition, object detection, instance segmentation and semantic segmentation without heavy supervised training. Moreover, we observe quantitative changes in scaling \eva result in qualitative changes in transfer learning performance that are not present in other models. For instance, \eva takes a great leap in the challenging large vocabulary instance segmentation task: our model achieves almost the same state-of-the-art performance on LVISv1.0 dataset with over a thousand categories and COCO dataset with only eighty categories. Beyond a pure vision encoder, \eva can also serve as a vision-centric, multi-modal pivot to connect images and text. We find initializing the vision tower of a giant CLIP from \eva can greatly stabilize the training and outperform the training from scratch counterpart with much fewer samples and less compute, providing a new direction for scaling up and accelerating the costly training of multi-modal foundation models. To facilitate future research, we release all the code and billion-scale models.
\end{abstract}

%#################################################
% Summary of EVA performance
%#################################################
\begin{table*}[h] 
\vspace{-15pt}
    \centering
    \tablestyle{1.5pt}{1.15}
    \begin{tabular}{l|lllllll|lll|ll}
        & \multicolumn{7}{c|}{\scriptsize{image \& video classification ({\faIcon{github}})}} & \multicolumn{3}{c|}{\scriptsize{object detection (det) \& instance segmentation (seg)}} & \multicolumn{2}{c}{\scriptsize{semantic segmentation}} \\
        % \hline
        model & {\scriptsize{IN-1K}} ft & {\scriptsize{IN-1K}} lin & {\scriptsize{IN-1K}} zs & avg. zs & \scriptsize{K400} & \scriptsize{K600} & \scriptsize{K700} & \scriptsize{COCO} det \scriptsize{($\mathtt{test}$/$\mathtt{val}$)} & \scriptsize{COCO} seg \scriptsize{($\mathtt{test}$/$\mathtt{val}$)} & \scriptsize{LVIS} seg & \scriptsize{COCO-Stuff} & \scriptsize{ADE20K} \\
        \shline
        Florence & \ph{---}- & \ph{---}- & \ph{---}- & \ph{---}- & 86.5 & 87.8 & \ph{---}- & 62.4\ph{\suptext{e}}/ 62.0 & \ph{62.4\suptext{e}}- \ph{62.0} & \ph{---}- & \ph{---}- & \ph{---}- \\
        SwinV2-G & \ph{---}- & \ph{---}- & \ph{---}- & \ph{---}- & 86.8 & \ph{---}- & \ph{---}- & 63.1\ph{\suptext{e}}/ 62.5 & 54.4\ph{\suptext{e}}/ 53.7 & \ph{---}- & \ph{---}- & 59.9 \\
        prev. best & 89.6\suptext{\demphs{a}} & 82.3\suptext{\demphs{b}} & 78.0\suptext{\demphs{c}} & 73.1\suptext{\demphs{c}} & 87.8\suptext{\demphs{d}} & 88.3\suptext{\demphs{e}} & 80.4\suptext{\demphs{e}} & 64.5\suptext{\demphs{f}}/ 64.2\suptext{\demphs{g}} & 55.4\suptext{\demphs{h}}/ 54.5\suptext{\demphs{i}} & 49.2\suptext{\demphs{j}} & 52.3\suptext{\demphs{k}} & 62.8\suptext{\demphs{a}} \\
        \eva & \textbf{89.7}\dtplus{+0.1} 
        & \textbf{86.5}\dtplus{+4.2} 
        & \textbf{78.5}\dtplus{+0.5} 
        & \textbf{75.7}\dtplus{+2.6} 
        & \textbf{89.7}\dtplus{+1.9} 
        & \textbf{89.8}\dtplus{+1.5} 
        & \textbf{82.9}\dtplus{+2.5} & \textbf{64.7}\ph{\suptext{e}}/ {\textbf{64.5}\ph{\suptext{}}\dtplus{+0.2/+0.3}} & {\textbf{55.5}\ph{\suptext{e}}}/ {\textbf{55.0}\ph{\suptext{}}\dtplus{+0.1/+0.5}} & \textbf{55.0}\dtplus{+5.8} & \textbf{53.4}\dtplus{+1.1} & \textbf{62.3}\demphs{\dt{-0.5}}
        \end{tabular}
        \vspace{-4pt}
        \caption{\textbf{Summary of \eva performance on various mainstream vision benchmarks.} \eva is performant compared with previous best / leading approaches. {\scriptsize ``\faIcon{github}''}: methods / results that only exploit publicly accessible data / academic resources. ``ft'': end-to-end fine-tuning. ``lin'': linear probing. ``zs'': zero-shot classification. ``avg. zs'': averaged zero-shot classification performance on 8 image and 4 video datasets with contrastive language-image pre-training. {\footnotesize{\demphinline{({timestamp: Nov 10, 2022})}}} \\ {\scriptsize methods / results reference. a: BEiT-3~\cite{beit3}, b: iBOT~\cite{zhou2021ibot}, c: Open CLIP-H~\cite{openclip}, d: Text4Vis~\cite{wu2022transferring}, e: MaskFeat~\cite{wei2022masked}, f: Group DETRv2~\cite{chen2022group}, g: FocalNet~\cite{yang2022focal}, h: FD-SwinV2-G~\cite{wei2022featdistill}, i: Mask DINO~\cite{maskdino}, j: LVIS 2021 competition 1\suptext{st}~\cite{fu2021lvis}, k: ViT-Adapter~\cite{vitadapt}.}}
        \vspace{-5pt}
        \label{tab: summary of eva performance}
\end{table*}

\section{Introduction}
\label{sec:intro}

Scaling up pre-trained language models (PLMs)~\cite{liu2019roberta,gpt3,t5} has revolutionized natural language processing (NLP) in the past few years.
The key to this success lies in the simple and scalable self-supervised learning task of masked signal prediction~\cite{radford2018improving,devlin2018bert}, with which Transformer models~\cite{vaswani2017attention} could be scaled up to billions of parameters using nearly unlimited unlabelled data, and generalize well to a wide range of downstream tasks with little tuning.
With further scaling on compute, data, and model sizes, PLMs have led to not only continuous performance improvements~\cite{t5,kaplan2020scalingLM,rae2021gopher}, but also a surprising emergence of in-context learning capability~\cite{gpt3,wei2021finetuned,chowdhery2022palm,wei2022emergent}.

\begin{figure}
    \centering
    \includegraphics[width=\linewidth]{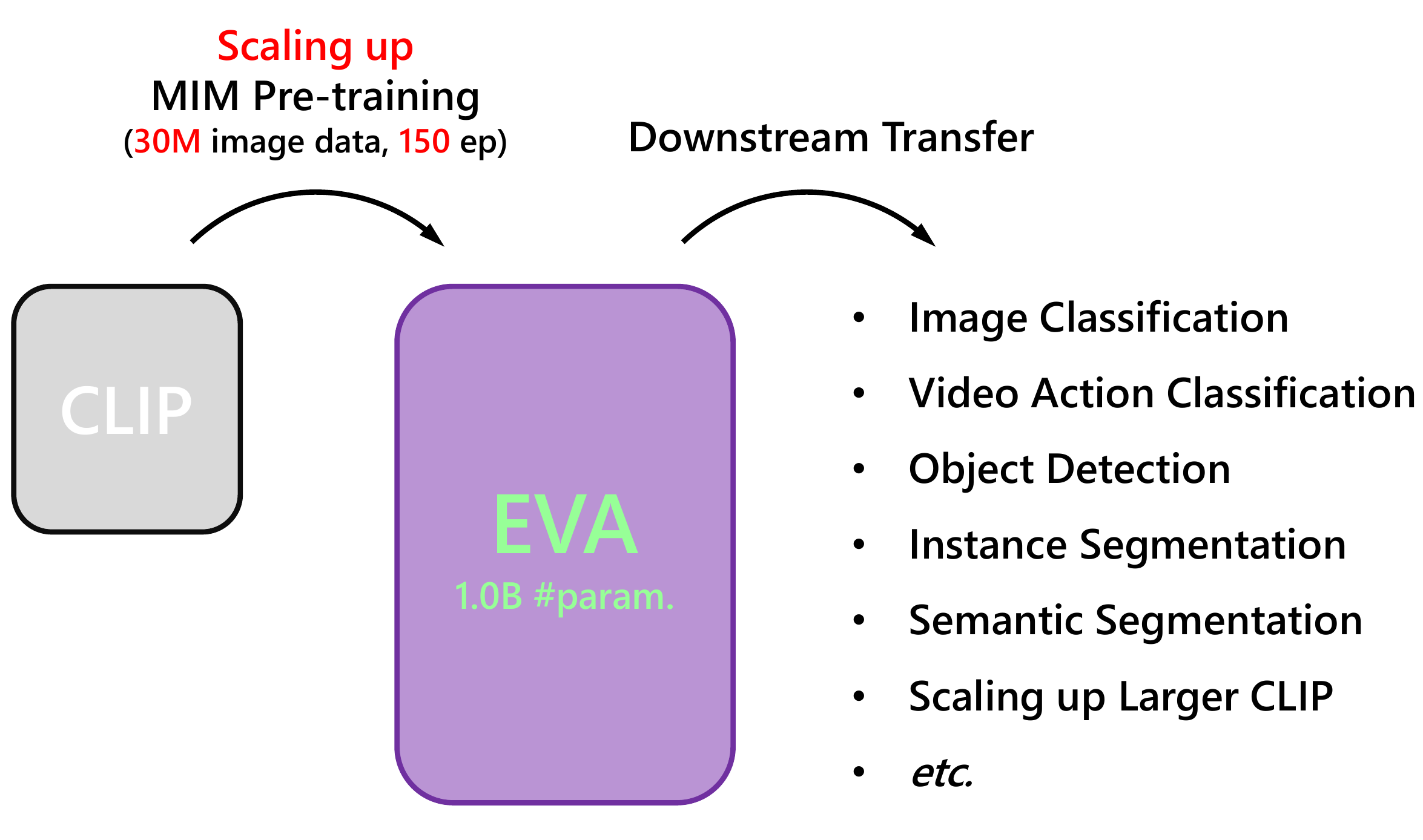}
    \caption{\textbf{An illustration of this work.} We find with sufficient image data (30M) and compute (150 epochs), simply regressing the masked out image-text aligned vision features (\ie, CLIP features) scales up well (to 1.0B parameters) and transfers well to various downstream tasks.}
    \label{fig:teaser}
\end{figure}

Motivated by the success of model scaling in NLP, it is appealing that we can also translate this success from language to vision, \ie, to scale up a vision-centric foundation model that is beneficial for both vision \& multi-modal downstream tasks.
Recently, masked image modeling (MIM)~\cite{bao2021beit,xie2021simmim,he2021masked} has boomed as a viable approach for vision model pre-training and scaling.
However, the most competitive billion-sized vision pre-trained models~\cite{dosovitskiy2020vit,zhai2022scalingvit,pham2021combined,swinv2} still heavily rely on supervised or weakly-supervised training with hundreds of millions of (often publicly inaccessible) labeled data.
MIM is somewhat only adopted as an initialization stage before the heavily supervised pre-training~\cite{swinv2}, or a pure MIM pre-trained model could not achieve favorable performance at billion-scale model sizes~\cite{xie2022datascaling}.
We regard this gap stems from the fact that natural images are raw and information-sparse.
Meanwhile, an ideal vision pretext task needs the abstraction of not only the low-level geometry \& structure information, but also high-level semantics, which is hardly captured by pixel-level recovery tasks~\cite{xie2022understandingmim}.

In this work, we seek a suitable MIM pretext task for large scale vision representation learning and explore its limits at the scale of one billion parameters with tens of millions of unlabeled data.
Recently, there are a few trials leveraging the semantic information from image-image or image-text contrastive learning~\cite{caron2021emerging,chen2021mocov3,clip} for MIM pre-training~\cite{zhou2021ibot,wei2022mvp,hou2022milan}, which perform fairly well in vision downstream tasks.
However, there remains a debate that 
\textbf{(i)} tokenized semantic features could provide better supervision signal for masked modeling in vision~\cite{bao2021beit, beitv2, beit3}, and 
\textbf{(ii)} good performances could be also achieved via a simple post-distillation process without masked prediction tasks~\cite{wei2022featdistill}.
Through a pilot empirical study, we find that simply using image-text aligned (\ie, CLIP~\cite{clip}) vision features as the prediction targets in MIM scales up well and achieves satisfactory performances on a broad range of downstream benchmarks.
This pre-training task draws the benefits from both the high-level semantic abstraction of image-text contrastive learning as well as the good capture of geometry \& structure in masked image modeling, which typically covers the information needed for most visual perception tasks.

%#################################################
% pilot exp
%#################################################
\begin{table}[!b]
% \vspace{-1em}
\centering
%#################################################
% tokenized?
%#################################################
\subfloat[
\textbf{(Additional) semantic feature tokenization} is not required for achieving good downstream performance.
\label{tab: tokenizer}
]{
\centering
\tablestyle{6pt}{1.05}
\begin{tabular}{x{40}x{40}|x{40}x{40}}
& & {\scriptsize{ImageNet-1K}} & {\scriptsize{ADE20K}} \\
tokenize?~\cite{beitv2} & pt epochs & top-1 acc. & \mIoUss \\
\shline
{\xmark} & {-} & \demphs{85.0} & \demphs{52.6} \\
\cmark & 300 & 85.0 & 52.7 \\
\cmark & 1600 & 85.5 & 53.1 \\
\evadefault{\xmark} & {800} & \bf\evagreen{85.5} & \bf\evagreen{53.3} \\
\end{tabular}
}
\vspace{.5em}
\\
%#################################################
% distillation?
%#################################################
\subfloat[
\textbf{Feature distillation} fails to achieve consistent performance gain as the pre-training becomes longer.
\label{tab: distill}
]{
\centering
\tablestyle{6pt}{1.05}
\begin{tabular}{x{40}x{40}|x{40}x{40}}
& & {\scriptsize{ImageNet-1K}} & {\scriptsize{ADE20K}} \\
distill.?~\cite{wei2022featdistill} & pt epochs & top-1 acc. & \mIoUss \\
\shline
\xmark & - & \demphs{85.0} & \demphs{52.6} \\
\cmark & 300 & 85.1 & 52.5 \\
\cmark & 800 & 85.1 & 52.7 \\
\evadefault{\xmark} & {800} & \bf\evagreen{85.5} & \bf\evagreen{53.3} \\
\end{tabular}
}
\caption{\textbf{Pilot experiment}. We evaluate different pre-training approaches using ViT-B and report their performance on ImageNet-1K image classification (top-1 accuracy) and ADE20K semantic segmentation (single-scale mIoU). \demphs{Numbers in grey} refer to the results of directly fine-tuning CLIP vision encoder on corresponding downstream tasks.  Default settings for \eva pre-training are marked in \colorbox{defaultcolor}{purple}, \ie, directly regressing the masked out CLIP vision features conditioned on visible image patches.}
\label{tab: pilot exp}
% \vspace{-1.em}
\end{table}

%#################################################
% eva pt cfg
%#################################################
\begin{table*}[!t]
% \vspace{-15pt}
\centering
%#################################################
% arch cfg
%#################################################
\subfloat[
\eva architecture configurations.
\label{tab: pt arch}
]{
\centering
\begin{minipage}{0.48\linewidth}{\begin{center}
\tablestyle{3.5pt}{1.2}
    \begin{tabular}{ccccc|c}
            patch size & \#layers & hidden dim & mlp dim & attn heads & \#param. \\
            \shline
            14$\times$14 & 40 & 1408 & 6144 & 16 & 1011M \\
    \end{tabular}
\end{center}}\end{minipage}
}
% \\
%#################################################
% pt data
%#################################################
\subfloat[
datasets for pre-training \eva.
\label{tab: pt data}
]{
\centering
\begin{minipage}{0.48\linewidth}{\begin{center}
    \tablestyle{2.0pt}{1.2}
    \begin{tabular}{cccccc|c}
            \multicolumn{6}{c|}{dataset} & total size \\
            \shline
            \scriptsize{ImageNet-21K,} & \scriptsize{CC12M,} & \scriptsize{CC3M,} & \scriptsize{Object365,} & \scriptsize{COCO,} & \scriptsize{ADE} & 29.6M images \\
    \end{tabular}
\end{center}}\end{minipage}
}
\vspace{0.5em}
\\
%#################################################
% pt setting & hyperparams
%#################################################
\subfloat[
some pre-training settings and hyper-parameters.
\label{tab: pt setting and hyperparams}
]{
\centering
\begin{minipage}{0.49\linewidth}{\begin{center}
    \tablestyle{3pt}{1.2}
    \begin{tabular}{cccccc}
            image size & batch size & optimizer & peak lr & ($\beta_1$, $\beta_2$) & pt epochs \\
            \shline
            224\suptext{2} & 4096 & AdamW & 1e-3 & (0.9, 0.98) & 150 \\
    \end{tabular}
\end{center}}\end{minipage}
}
%#################################################
% pt stat
%#################################################
\subfloat[
basic statistics of \eva pre-training.
\label{tab: pt stat}
]{
\centering
\begin{minipage}{0.49\linewidth}{\begin{center}
    \tablestyle{4.5pt}{1.2}
    \begin{tabular}{cccccc}
            precision & ZeRO & \#gpus & samples / sec. & max mem. & pt days \\
            \shline
            $\mathtt{fp16}$ & stage-1 & 128 & \app3150 & \app26.5GB & \app14.5 \\
    \end{tabular}
\end{center}}\end{minipage}
}
\\
\vspace{-0.5em}
\caption{\bf{A brief summary of pre-training settings and configurations for \eva.}}
\label{tab: pt cfg}
\vspace{-0.5em}
\end{table*}

Via this MIM pretext task, we can efficiently scale up a vanilla ViT encoder~\cite{dosovitskiy2020vit}, dubbed \eva, to one billion parameters with strong visual representations that transfers well to a wide range of downstream tasks (\figref{fig:teaser}).
Using 29.6 million public accessible unlabeled images for pre-training, \eva sets new records on several representative vision benchmarks, such as image classification on ImageNet-1K~\cite{deng2009imagenet} {\footnotesize{\demphinline{(89.7\% top-1 accuracy)}}}, object detection and instance segmentation on LVISv1.0~\cite{gupta2019lvis} {\footnotesize{\demphinline{(62.2 \boxAP \& 55.0 \maskAP on \val)}}} and COCO~\cite{lin2014coco} {\footnotesize{\demphinline{(64.5 \boxAP \& 55.0 \maskAP on \val, 64.7 \boxAP \& 55.5 \maskAP on \test)}}}, semantic segmentation on COCO-stuff~\cite{coco-stuff} {\footnotesize{\demphinline{(53.4 \mIoUss)}}} and ADE20K~\cite{zhou2018ade} {\footnotesize{\demphinline{(62.3 \mIoUms)}}}, and video action recognition on Kinetics-400~\cite{kay2017kinetics} {\footnotesize{\demphinline{(89.7\% top-1 accuracy)}}}, Kinetics-600~\cite{k600} {\footnotesize{\demphinline{(89.8\% top-1 accuracy)}}}, Kinetics-700~\cite{k700} {\footnotesize{\demphinline{(82.9\% top-1 accuracy)}}}.
Notably, different from other state-of-the-art billion-scale vision foundation models that demand tens of millions of or even billions of labeled images, such as SwinV2-G using ImageNet-21K-ext-70M~\cite{swinv2} and ViT-g/G using JFT-3B~\cite{zhai2022scalingvit}, \eva does not need a costly supervised training stage and only leverage images from open-sourced datasets for academic reproducibility.

Moreover, we observe quantitative changes in scaling \eva result in qualitative changes in transfer learning performance that are not observed in other smaller-scale models, \eg, \eva makes a significant breakthrough in the challenging large vocabulary object-level recognition task: our model achieves almost the same performance on LVISv1.0~\cite{gupta2019lvis} {\footnotesize{\demphinline{(55.0 \maskAP on \val)}}}, an instance segmentation benchmark with more than 1,200 categories, as COCO~\cite{lin2014coco} {\footnotesize{\demphinline{(55.0 \maskAP on \val)}}}, which almost shares the same image set as LVISv1.0 but with only 80 categories annotated.
This emergent ability well matches the expectation of model scaling~\cite{wei2022emergent}, that larger capability of model results in not only predictable performance improvements on standard benchmarks, but also unpredictable phenomenons and capabilities for resolving more challenging tasks.

Going beyond a pure vision encoder, \eva can also serve as a vision-centric, multi-modal pivot that builds a bridge between vision and language.
We show that initializing the image encoder via pre-trained \eva in a 1.1 billion parameters CLIP model can outperform the training from scratch counterpart on a broad range of zero-shot image / video classification benchmarks with much fewer samples and less compute.
Moreover, \eva can greatly stabilize the giant CLIP's training \& optimization process.
Since large CLIP models usually suffer from training instability and inefficiency issues~\cite{openclip, laiongiantclip}, we hope our solution opens up a new direction for scaling up and accelerating the costly training of multi-modal foundation models.

By scaling up vision-centric foundation models with MIM pre-training to achieve strong performance on broad downstream tasks, we hope \eva would bridge the gap between vision and language with masked signal modeling, and contributes to the big convergence across different modalities.

\section{Fly \eva to the Moon}
\label{sec: exp}
We first conduct a series of pilot experiments for choosing an ideal vision pretext task in \sref{sec:approach}, then we scale up \eva pre-training via the chosen pre-training objective in \sref{sec: pt exp}.
Finally, we evaluate the pre-trained representation on various downstream tasks in \sref{sec: downstream tasks}.
Detailed experimental settings and configurations are in Appendix~\ref{app}.

\subsection{The Feature Instrumentality Project}
\label{sec:approach}

In this section, we seek a MIM vision pretext task with compelling transfer performance.
Based on previous literature on vision pre-training, we study two promising candidates: \textbf{(i)} recovering the masked out \textit{tokenized} semantic vision features~\cite{bao2021beit,beitv2,beit3}, and 
\textbf{(ii)} feature \textit{distillation} from strong pre-trained representation as in~\cite{wei2022featdistill}.
Both of them exploit pre-trained image-text aligned vision features (\ie, CLIP~\cite{clip} vision features).
Via a series of pilot experiments shown in \tblref{tab: pilot exp}, we find that:
\textbf{(i)} the (additional) CLIP feature tokenization process is unnecessary for achieving good downstream performance
\textbf{(ii)} feature \textit{distillation} fails to provide consistent performance gain as the pre-training becomes longer.
Instead, we find that simply reconstructing the masked out CLIP vision features conditioned on visible image patches is highly performant, which is chosen for scaling up \eva. 

We clarify that this MIM pretext task is \textit{not} originally proposed by us.
Regressing the masked out image-text aligned vision features for MIM pre-training has been studied in MVP~\cite{wei2022mvp} and recently has been revisited by MILAN~\cite{hou2022milan}.
In this work, we show that this pretext task can scale up to billion-scale parameters and tens of millions of unlabeled images for vision-centric representation learning \textit{without} \textbf{(i)} semantic feature quantization / tokenization~\cite{bao2021beit,beitv2}, and \textbf{(ii)} explicitly using image-text paired pre-training data and large corpora as in BEiT-3~\cite{beit3}.

\subsection{Pre-training}
\label{sec: pt exp}

\paragraph{Architecture.}
The architecture configurations of \eva are in \tblref{tab: pt arch}.
\eva is a vanilla ViT~\cite{dosovitskiy2020vit} with 1.0B parameters. 
The shape of her follows ViT giant~\cite{zhai2022scalingvit} and the vision encoder of BEiT-3~\cite{beit3}. We do not use relative positional embeddings~\cite{rpe} and layer-scale~\cite{cait} during pre-training.

\paragraph{Pre-training objective.}
\eva is pre-trained to reconstruct the masked out image-text aligned vision features conditioned on visible image patches.
We corrupt the input patches with $\mathtt{[MASK]}$ tokens, and we use block-wise masking with a masking ratio of 40\% following~\cite{bao2021beit, beitv2, beit3}.
The target for MIM pre-training is from the publicly available\footnote{Source: \url{https://github.com/openai/CLIP}} OpenAI CLIP-L/14 vision tower trained on 224$\times$224 pixel images~\cite{clip}.
The output feature of \eva is first normalized~\cite{ln} and then projected to the same dimension as the CLIP feature via a linear layer.
We use negative cosine similarity as the loss function.

\paragraph{Pre-training data.}
The data we used for pre-training \eva are summarized in \tblref{tab: pt data}.
For CC12M~\cite{CC12M} and CC3M~\cite{CC3M} datasets, we only use the image data without captions.
For COCO~\cite{lin2014coco} and ADE20K~\cite{zhou2018ade} datasets, we only use the train set data.
ImageNet-21K~\cite{deng2009imagenet} and Object365~\cite{o365} image data are also used.
All these data are publicly accessible.
The merged dataset for pre-training has 29.6 million images in total.

The CLIP features we used as MIM prediction targets are trained on a 400 million image-text dataset in a self-supervised manner.
So during pre-training \eva also implicitly exploits the knowledge from this dataset to some extent.
Meanwhile, these CLIP features is also widely used in other state-of-the-art representation learning \& pre-training works such as the BEiT family~\cite{beitv2,beit3}, AI generated content~\cite{dalle2,imagen,stablediffusion} and large scale dataset filtering~\cite{kakaobrain2022coyo-700m,laion400m,laion5b}.

\paragraph{Pre-training settings \& hyper-parameters.}
As shown in \tblref{tab: pt setting and hyperparams}, \eva is optimized via Adam~\cite{adam} with decoupled weight decay~\cite{Loshchilov2019adamw} of 0.05.
The peak learning rate is 1e-3 and decays according to a cosine learning rate schedule.
We employed stochastic depth~\cite{huang2016deep} with a rate of 0.1 for regularization and $\mathtt{RandResizeCrop}$ (0.2, 1) for data augmentation. 
Color jitter is not used.

\paragraph{Pre-training infrastructure and statistics.}
Some basic pre-training statistics are available in \tblref{tab: pt stat}. 
The GPU we use is NVIDIA A100-SXM4-40GB. 
Pre-training code is based on BEiT~\cite{bao2021beit} written in PyTorch~\cite{pytorch}.
We also adopt $\mathtt{DeepSpeed}$ optimization library~\cite{rasley2020deepspeed} with ZeRO stage-1 optimizer~\cite{rajbhandari2020zero} to save memory.
We find using $\mathtt{fp16}$ format with dynamic loss scaling is stable enough during the whole course of pre-training while using $\mathtt{bfloat16}$ format is unnecessary.
Since we use $\mathtt{fp16}$ precision, \eva can also be pre-trained using 16$\times$ NVIDIA 24GB (32GB) GPUs with (without) gradient checkpointing~\cite{gradckpt}.

\subsection{Evaluation on Downstream Tasks}
\label{sec: downstream tasks}
In this section, we extensively evaluate pre-trained \eva on several representative benchmarks, such as image classification (\sref{sec: image classification}), video action recognition (\sref{sec: video}), object detection \& instance segmentation (\sref{sec: object detection}), semantic segmentation (\sref{sec: semantic segmentation}), and contrastive image-text pre-training with zero-shot evaluation (\sref{sec: clip exp}).
\eva achieves state-of-the-art performance on a broad range of downstream tasks.

%#################################################
% image classification
%#################################################
\begin{table}[!t]
% \vspace{-10pt}
    \centering
    \tablestyle{2.5pt}{1.15}
    \begin{tabular}{l|ccc|c}
            model & \#param. & extra labeled data & image size & top-1 acc. \\
            \shline
            \multicolumn{5}{c}{\scriptsize{\demphs{using \textit{private} labeled data}}} \\
            \hline
            \demphs{SwinV2-G}~\cite{swinv2} & \demphs{3.0B} & \demphs{\scriptsize{IN-21K-ext-70M}} & \demphs{640\suptext{2}} & \demphs{90.2} \\
            \demphs{ViT-G}~\cite{zhai2022scalingvit} & \demphs{1.8B} & \demphs{JFT-3B} & \demphs{518\suptext{2}} & \demphs{90.5} \\
            \demphs{ViT-g (CoCa)}~\cite{yu2022coca} & \demphs{1.0B} & \demphs{\scriptsize{JFT-3B+ALIGN}} & \demphs{576\suptext{2}} & \demphs{91.0} \\
            \shline
            \multicolumn{5}{c}{\scriptsize{using \textit{public} labeled data}} \\
            \hline
            CoAtNet-4~\cite{dai2021coatnet} & 275M & \scriptsize{IN-21K (14M)} & 512\suptext{2} & 88.6 \\
            MaxViT-XL~\cite{tu2022maxvit} & 475M & \scriptsize{IN-21K (14M)} & 512\suptext{2} & 88.7 \\
            MViTv2-H~\cite{li2021improved} & 667M & \scriptsize{IN-21K (14M)} & 512\suptext{2} & 88.8 \\
            FD-CLIP-L~\cite{wei2022featdistill} & 304M & \scriptsize{IN-21K (14M)} & 336\suptext{2} & 89.0 \\
            BEiT-3~\cite{beit3} & 2.0B & \scriptsize{35M img-txt pairs} & 336\suptext{2} & 89.6 \\
            \bf\evapurple\eva & 1.0B & \scriptsize{IN-21K (14M)} & 336\suptext{2} & \bf\evagreen{89.6} \\
            \bf\evapurple\eva & 1.0B & \scriptsize{IN-21K (14M)} & 560\suptext{2} & \bf\evagreen{89.7} \\
        \end{tabular}
        \vspace{-8pt}
        \caption{\textbf{Comparisons of image classification performance on ImageNet-1K validation set.} With only publicly available data, \eva creates a new state-of-the-art ImageNet-1K image classification result with a canonical linear classifier.}
        \vspace{-10pt}
        \label{tab: 1k cls}
\end{table}

\subsubsection{Image Classification}
\label{sec: image classification}

\paragraph{Datasets.}
For image classification task, we evaluate \eva on ImageNet-1K {\footnotesize{\demphinline{(IN-1K)}}}~\cite{deng2009imagenet} validation set.
We also evaluate the robustness \& generalization capability of \eva along with our training settings \& hyper-parameters using ImageNet-V2 matched frequency {\footnotesize{\demphinline{(IN-V2)}}}~\cite{inv2}, ImageNet-ReaL {\footnotesize{\demphinline{(IN-ReaL)}}}~\cite{inreal}, ImageNet-Adversarial {\footnotesize{\demphinline{(IN-Adv.)}}}~\cite{inadv}, ImageNet-Rendition {\footnotesize{\demphinline{(IN-Ren.)}}}~\cite{inren}, ImageNet-Sketch {\footnotesize{\demphinline{(IN-Ske.)}}}~\cite{inske}.

\paragraph{Training Settings.}
Following the conventional setting~\cite{bao2021beit,beitv2,beit3}, we first perform intermediate fine-tuning on ImageNet-21K~\cite{deng2009imagenet} for 60 epochs with an image resolution of 224\suptext{2}, then \eva is further fine-tuned on ImageNet-1K training set for 10 epochs.
Different from~\cite{zhai2022scalingvit,yu2022coca} that use multi-head attention pooling and BEiT-3 that exploits an additional pre-trained giant language tower as the image classification task layer, we simply adopt a linear layer as the classifier~\cite{dosovitskiy2020vit}.
Notice that the supervised intermediate fine-tuning consumes only \app1/5 of the time \& compute of the MIM pre-training stage.
While for other billion-scale vision models such as SwinV2-G-3B, the supervised training phase costs \app1.5$\times$ resources than the MIM pre-training.

\paragraph{Results.}
\tblref{tab: 1k cls} compares \eva with some state-of-the-art models on ImageNet-1K validation set.
\eva achieves 89.6\% top-1 accuracy with 336\suptext{2} inputs, comparable to BEiT-3.
Using a larger image resolution of 560\suptext{2} can further boost the top-1 accuracy to 89.7\%.
Notice that BEiT-3 treats image classification as an image-to-text retrieval task. 
Therefore they leverage an additional one billion parameters pre-trained language encoder along with 35 million image-text data {\footnotesize{\demphinline{(21M pairs from CC12M, CC3M, SBU, COCO, VG and 14M pairs from ImageNet-21K)}}} as well as 160GB text data in total.
Meanwhile, we simply use a linear classifier on top of \eva with only ImageNet-21K image-tag data used for additional fine-tuning.
With only publicly available data, \eva creates a new state-of-the-art image classification result on ImageNet-1K with a much neater architecture.

\paragraph{Robustness \& generalization ability evaluation.}
We evaluate the robustness and generalization capability of \eva trained with an image size of 336\suptext{2} on 6 different ImageNet-1K validation set variants.
In \tblref{tab: cls rob & gen}, we compare \eva with some top open-sourced models collected by the $\mathtt{timm}$ library\footnote{Source: \url{https://github.com/rwightman/pytorch-image-models/tree/main/results} {\footnotesize{\demphinline{({timestamp: Nov 10, 2022})}}}. The detailed model configurations are (arch-model\_size-img\_resolution-data): ConvNeXt-XL-384px-21K~\cite{convnext}, SwinV2-L-384px-21K~\cite{swinv2}, MAE-H-448px-1K~\cite{he2021masked}, DeiT3-L-384px-21K~\cite{deit3}, EfficientNet-L2\&NS-800px-JFT300M~\cite{ns}, BEiTv2-L-224px-21K~\cite{beitv2}, BEiT-L-512px-21K~\cite{bao2021beit}, \eva-g-336px-merged30M\&21k.}~\cite{rw2019timm}.
Following the evaluation procedure in~\cite{he2021masked}, all these models are first fine-tuned on the original ImageNet-1K training set and then evaluated on different validation sets using the \textit{same} fine-tuned model without further hyper-parameter selection and specialized fine-tuning.

As shown in \tblref{tab: cls rob & gen}, \eva is the most competitive one in terms of absolute top-1 accuracies.
However, these model various in pre-train data {\footnotesize{\demphinline{(from ImageNet-1K, ImageNet-21K to JFT-300M)}}}, input resolutions {\footnotesize{\demphinline{(from 224\suptext{2} to 800\suptext{2})}}}, model sizes {\footnotesize{\demphinline{(from hundreds of millions to one billion parameters)}}} as well as architectures {\footnotesize{\demphinline{(ConvNets, vanilla \& hierarchical ViTs)}}}, \etc.
Therefore their absolute accuracies are \textit{not} directly comparable.
Instead, we are more interested in the \textit{gap} between the averaged top-1 accuracy on 6 validation set variants and the original ImageNet-1K validation set top-1 accuracy (the lower the better), \ie, we care about whether a model along with its training settings biases towards the original validation set and generalize well on other variants.
From this perspective, \eva not only achieves the highest averaged accuracy, but also has the smallest performance gap, which reflects the excellent robustness and generalization ability of \eva.

%#################################################
% image classification
%#################################################
\begin{table}[!t] 
% \vspace{-10pt}
    \centering
    \tablestyle{2.8pt}{1.15}
    \begin{tabular}{l|cccccc|c|r}
        model & \scriptsize{\demphs{IN-1K}} & \scriptsize{\demphs{IN-V2}} & \scriptsize{\demphs{IN-ReaL}} & \scriptsize{\demphs{IN-Adv.}} & \scriptsize{\demphs{IN-Ren.}} & \scriptsize{\demphs{IN-Ske.}} & avg. & $\Delta$\scriptsize{$\downarrow$} \\
        \shline
        \scriptsize{ConvNeXt} & \scriptsize{\demphs{87.5}} & \scriptsize{\demphs{77.7}} & \scriptsize{\demphs{90.5}} & \scriptsize{\demphs{70.8}} & \scriptsize{\demphs{67.0}} & \scriptsize{\demphs{53.7}} & 74.5 & 13.0 \\
        SwinV2 & \scriptsize{\demphs{87.5}} & \scriptsize{\demphs{77.3}} & \scriptsize{\demphs{90.2}} & \scriptsize{\demphs{73.9}} & \scriptsize{\demphs{67.7}} & \scriptsize{\demphs{52.3}} & 74.8 & 12.7 \\
        MAE & \scriptsize{\demphs{87.8}} & \scriptsize{\demphs{79.2}} & \scriptsize{\demphs{90.3}} & \scriptsize{\demphs{76.7}} & \scriptsize{\demphs{66.5}} & \scriptsize{\demphs{50.9}} & 75.2 & 12.6 \\
        DeiT3 & \scriptsize{\demphs{87.7}} & \scriptsize{\demphs{79.1}} & \scriptsize{\demphs{90.2}} & \scriptsize{\demphs{79.2}} & \scriptsize{\demphs{70.6}} & \scriptsize{\demphs{54.9}} & 77.0 & 10.7 \\
        \scriptsize{Eff-L2-NS} & \scriptsize{\demphs{88.4}} & \scriptsize{\demphs{80.5}} & \scriptsize{\demphs{90.6}} & \scriptsize{\demphs{84.8}} & \scriptsize{\demphs{74.7}} & \scriptsize{\demphs{47.6}} & 77.8 & 10.6 \\
        BEiTv2 & \scriptsize{\demphs{88.4}} & \scriptsize{\demphs{80.1}} & \scriptsize{\demphs{90.3}} & \scriptsize{\demphs{76.2}} & \scriptsize{\demphs{76.4}} & \scriptsize{\demphs{58.3}} & 78.3 & 10.1 \\
        BEiT & \scriptsize{\demphs{88.6}} & \scriptsize{\demphs{79.9}} & \scriptsize{\demphs{90.7}} & \scriptsize{\demphs{81.7}} & \scriptsize{\demphs{73.2}} & \scriptsize{\demphs{56.8}} & 78.5 & 10.1 \\
        % \eva & \scriptsize{\demphs{89.6}} & \scriptsize{\demphs{81.1}} & \scriptsize{\demphs{90.6}} & \scriptsize{\demphs{84.9}} & \scriptsize{\demphs{87.3}} & \scriptsize{\demphs{66.8}} & 83.4 & \bf\evagreen{6.2} \\  bug: eval using non-ema ckpt
        \eva & \scriptsize{\demphs{89.6}} & \scriptsize{\demphs{81.6}} & \scriptsize{\demphs{90.8}} & \scriptsize{\demphs{86.2}} & \scriptsize{\demphs{88.3}} & \scriptsize{\demphs{67.7}} & 84.0 & \bf\evagreen{5.6} \\ % fix the bug, improve all results
        % we also test the objectnet top-1 acc.: 60.9, see our repo for details.
        \end{tabular}
        \vspace{-8pt}
        \caption{\textbf{Robustness \& generalization capability evaluation on ImageNet-1K variants.} We test each model on different ImageNet-1K validation sets, without any specialized fine-tuning. ``avg.'': the averaged top-1 accuracy on 6 different ImageNet-1K validation set variants. ``{$\Delta$\scriptsize{$\downarrow$}}'': The gap between the averaged top-1 accuracy on 6 variants ({\footnotesize{\demphinline{\ie, IN-\{1K, V2, ReaL, Adv., Ren., Ske.\}}}}) and the original ImageNet-1K validation set top-1 accuracy (the lower the better).}
        \vspace{-5pt}
        \label{tab: cls rob & gen}
\end{table}

\subsubsection{Video Action Recognition}
\label{sec: video}

\paragraph{Datasets.}
For video action recognition, we evaluate \eva on Kinetics-400 (K-400)~\cite{kay2017kinetics}, Kinetics-600 (K-600)~\cite{k600} and Kinetics-700 (K-700)~\cite{k700} benchmarks.
We first conduct intermediate fine-tuning on a merged dataset coined Kinetics-722 (K-722) that integrates videos from K-400, K-600 and K-700.
We remove leaked as well as repeated videos in both training and validation sets.
After this data de-duplicating process, K-722 has 0.63M training videos in total with 722 action classes.

\paragraph{Training \& evaluation settings.}
\eva processes video data simply via spatial-temporal attention as~\cite{feichtenhofer2022masked,tong2022videomae} with no specific architectural adaptation for video related tasks.
We first train \eva using K-722 training set for 40 epochs with 8 frames and 224\suptext{2} resolution, then we fine-tune \eva on each dataset for only 1 or 2 epochs.
We set frame$\times$crop$\times$clip to 16$\times$3$\times$4 for fine-tuning and evaluation for all datasets.
The frame resolution is 224\suptext{2}.

\paragraph{Results.}
As shown in \tblref{tab: video recognition}, \eva achieves better performance compared with some recent video-specific or large foundation models in video recognition.
For reference, directly adapting image-only pre-trained \eva to K-400 without K-722 intermediate fine-tuning can also achieve a very competitive top-1 accuracy of 88.4\%.

%#################################################
% video action recognition
%#################################################
\begin{table}[!t] 
    \centering
    \vspace{-10pt}
    \tablestyle{6pt}{1.1}
    \begin{tabular}{l|ccc}
        & \multicolumn{3}{c}{\scriptsize{top-1 accuracy}} \\
        model & \scriptsize{Kinetics-400} & \scriptsize{Kinetics-600} & \scriptsize{Kinetics-700} \\
        \shline
        MAE~\cite{feichtenhofer2022masked} & 86.8 & - & - \\
        SwinV2-G~\cite{swinv2} & 86.8 & - & - \\
        Florence~\cite{yuan2021florence} & 86.8 & 88.0 & - \\
        MaskFeat~\cite{wei2022masked} & 87.0 & 88.3 & 80.4 \\
        VideoMAE~\cite{tong2022videomae} & 87.4 & - & - \\
        X-CLIP~\cite{xclip} & 87.7 & 88.3 & - \\
        CoVeR~\cite{cover} & 87.2 & 87.9 & 78.5 \\
        CoCa~\cite{yu2022coca} (frozen) & 88.0 & 88.5 & 81.1 \\
        CoCa~\cite{yu2022coca} (finetuned) & 88.9 & 89.4 & 82.7 \\
        \bf\evapurple{\eva} & \bf\evagreen{89.7} & \bf\evagreen{89.8} & \bf\evagreen{82.9} \\
        \end{tabular}
        \vspace{-8pt}
        \caption{\textbf{Video action recognition.} With only publicly available K-400, K-600 and K-700 as video pre-training data, \eva is also quite performant in video action recognition tasks.}
        \label{tab: video recognition}
\vspace{-8pt}
\end{table}

%#################################################
% object detection (2 col)
%#################################################
\begin{table*}[!t] 
\vspace{-22pt}
\centering
\tablestyle{5.2pt}{1.15}
    \begin{tabular}{l|c|r|cc|c|cc|cc}
    & & & \multicolumn{2}{c|}{\demphs{pre-training data}} & & \multicolumn{2}{c|}{\scriptsize{COCO $\mathtt{val}$}} & \multicolumn{2}{c}{\scriptsize{COCO $\mathtt{test}$-$\mathtt{dev}$}} \\
    model / method & detector & \demphs{\#param.\ph{-}} & \demphs{encoder} & \demphs{detector} & tta? & \boxAP & \maskAP & \boxAP & \maskAP \\
    \shline
    Soft-Teacher~\cite{softteacher} & HTC++~\cite{chen2019htc} & \demphs{284M} & \demphs{IN-21K (14M)} & \demphs{\scriptsize{COCO(unlabeled)+O365}} & \cmark & 60.7 & 52.5 & 61.3 & 53.0 \\
    GLIP~\cite{glip} & DyHead~\cite{dai2021dynamic} & \demphs{${\scriptstyle \mathtt{\geq}}$ {284M}} & \demphs{IN-21K (14M)} & \demphs{\scriptsize{4ODs+GoldG+Cap12M}} & \cmark & 60.8 & - & 61.5 & - \\
    GLIPv2~\cite{glipv2} & DyHead~\cite{dai2021dynamic} & \demphs{\demphs{${\scriptstyle \mathtt{\geq}}$ 637M}} & \demphs{FLD-900M} & \demphs{merged data\suptext{a}} & \cmark & - & - & 62.4 & - \\
    ViTDet-H~\cite{li2022exploring} & CMask R-CNN~\cite{cai2019cascade} & \demphs{692M} & \demphs{IN-1K (1M)} & \demphs{-} & \cmark & 61.3 & 53.1 & - & - \\
    Florence~\cite{yuan2021florence} & DyHead~\cite{dai2021dynamic} & \demphs{${\scriptstyle \mathtt{\geq}}$ 637M} & \demphs{FLD-900M} & \demphs{merged data\suptext{a}} & \cmark & 62.0 & - & 62.4 & - \\
    SwinV2-G~\cite{swinv2} & HTC++~\cite{chen2019htc} & \demphs{${\scriptstyle \mathtt{\geq}}$ 3000M} & \demphs{IN-21K-ext-70M} & \demphs{O365} & \cmark & 62.5 & 53.7 & 63.1 & 54.4 \\
    DINO~\cite{detdino} & - & \demphs{218M} & \demphs{IN-21K (14M)} & \demphs{O365} & \cmark & 63.2 & - & 63.3 & - \\
    Mask DINO~\cite{maskdino} & - & \demphs{223M} & \demphs{IN-21K (14M)} & \demphs{O365} & \cmark & - & 54.5 & - & 54.7 \\
    BEiT-3~\cite{beit3} & {CMask R-CNN}~\cite{cai2019cascade} & \demphs{1074M} & \demphs{merged data\suptext{b}} & \demphs{O365} & \cmark & - & - & 63.7 & 54.8 \\
    FD-SwinV2-G~\cite{wei2022featdistill} & HTC++~\cite{chen2019htc} & \demphs{${\scriptstyle \mathtt{\geq}}$ 3000M} & \demphs{IN-21K-ext-70M} & \demphs{O365} & \cmark & - & - & 64.2 & 55.4 \\
    FocalNet~\cite{yang2022focal} & DINO~\cite{detdino} & \demphs{746M} & \demphs{IN-21K (14M)} & \demphs{O365} & \cmark & 64.2 & - & 64.4 & - \\
    Group DETRv2~\cite{chen2022group} & DINO~\cite{detdino} & \demphs{629M} & \demphs{IN-1K (1M)} & \demphs{O365} & \cmark & - & - & 64.5 & - \\
    \bf\evapurple{\eva} & {CMask R-CNN}~\cite{cai2019cascade} & \demphs{1074M} & \demphs{merged-30M} & \demphs{O365} & \xmark & \bf\evagreen{64.2} & \bf\evagreen{55.0} & \bf\evagreen{64.4} & \bf\evagreen{55.5} \\
    \bf\evapurple{\eva} & {CMask R-CNN}~\cite{cai2019cascade} & \demphs{1074M} & \demphs{merged-30M} & \demphs{O365} & \cmark & \bf\evagreen{64.5} & - & \bf\evagreen{64.7} & -
    \end{tabular}
\vspace{-.3em}
\caption{\textbf{Object detection \& instance segmentation on results COCO dataset.} \eva establishes new state-of-the-art results in object detection and instance segmentation tasks on both COCO $\mathtt{val}$ and $\mathtt{test}$-$\mathtt{dev}$ splits with the canonical R-CNN~\cite{rcnn} object detection \& segmentation framework. ``tta'' refers to test-time augmentation. {\footnotesize{\demphinline{({timestamp: Nov 10, 2022})}}} \\ {\scriptsize \ph{+} ``merged data\suptext{a}'': FourODs + INBoxes + GoldG + CC15M + SBU, \ph{+} ``merged data\suptext{b}'': IN-21K (14M) + Image-Text (35M) + Text (160GB).}}
\label{tab: who he sota}
\end{table*}

%#################################################
% object detection
%#################################################
\begin{table*}[t!]
% \vspace{-.4em}
\centering
%#################################################
% COCO
%#################################################
\subfloat[
\textbf{Results of object detection and instance segmentation on LVISv1.0 $\mathtt{val}$ set.} ``\boxAP, \maskAP, \maskAPrare'' are evaluated using the rule defined in~\cite{gupta2019lvis}. We also report the experimental ``boundary, fixed AP''(\demphs{\boundaryAPfix}) used in the LVIS 2021 challenge for reference.
\label{tab: lvis}
]{
\centering
\begin{minipage}{0.46\linewidth}{\begin{center}
\tablestyle{3.5pt}{1.1}
\begin{tabular}{l|ccc|c}
 &  &  &  &  \\
model & \boxAP & \maskAP & \maskAPrare & \demphs{\boundaryAPfix} \\
\shline
2020 competition 1\suptext{st}~\cite{tan20201st} & - & 41.5 & - & \demphs{-} \\
2021 competition 1\suptext{st}~\cite{fu2021lvis} & - & 49.2 & 45.4 & \demphs{44.1} \\
ViTDet-H~\cite{li2022exploring} & 53.4 & 48.1 & 36.9 & - \\
\textbf{\evapurple{\eva}} \scriptsize{(single-scale test)} & \bf\evagreen{62.2} & \bf\evagreen{55.0} & \bf\evagreen{48.3} & \bf\demphs{48.3} \\
\end{tabular}
\end{center}}\end{minipage}
}
\hspace{2em}
%#################################################
% LVIS
%#################################################
\subfloat[
\textbf{LVISv1.0 \& COCO performance gap on $\mathtt{val}$ set.} ``prev. best'' refers to the best \textit{individual} model / result in each benchmark (a: DINO~\cite{detdino}, b: ViTDet-H~\cite{li2022exploring}, c: Mask DINO~\cite{maskdino}, d: 2021 competition 1\suptext{st}~\cite{fu2021lvis}) ``{$\Delta$\scriptsize{$\downarrow$}}'': the performance gap between LVIS and COCO (the lower the better).
\label{tab: lvis coco gap}
]{
\centering
\begin{minipage}{0.46\linewidth}{\begin{center}
    \tablestyle{3.6pt}{1.1}
    \begin{tabular}{l|ccr|ccr}
        & \multicolumn{3}{c|}{\boxAP} & \multicolumn{3}{c}{\maskAP} \\
        model & COCO & LVIS & $\Delta$\scriptsize{$\downarrow$} & COCO & LVIS & $\Delta$\scriptsize{$\downarrow$} \\
        \shline
        {Copy-Paste~\cite{simple_copy_paste}} & 57.0\ph{\suptext{a}} & 41.6\ph{\suptext{a}} & 15.4 & 48.9\ph{\suptext{a}} & 38.1\ph{\suptext{a}} & 10.8 \\
        ViTDet-H~\cite{li2022exploring} & 61.3\ph{\suptext{a}} & 53.4\ph{\suptext{a}} & 7.9 & 53.1\ph{\suptext{a}} & 48.1\ph{\suptext{a}} & 5.0 \\
        prev. best & 63.2\suptext{\demphs{a}} & 53.4\suptext{\demphs{b}} & 9.8 & 54.5\suptext{\demphs{c}} & 49.2\suptext{\demphs{d}} & 5.3 \\
        \textbf{\evapurple{\eva}} \scriptsize{(single-scale test)} & \textbf{64.1}\ph{\suptext{a}} & \textbf{62.2}\ph{\suptext{a}} & \bf\evagreen{1.9} & \evadefault{\textbf{55.0}\textcolor{defaultcolor}{\suptext{a}}} & \evadefault{\textbf{55.0}\textcolor{defaultcolor}{\suptext{a}}} & \evadefault{\bf\evagreen{0.0}} \\
        \end{tabular}
\end{center}}\end{minipage}
}
\\
\vspace{-.3em}
\caption{\textbf{Object detection and instance segmentation performance on LVISv1.0 dataset} (\tblref{tab: lvis}). \eva not only achieves the state-of-the-art results on both LVISv1.0 and COCO with the same set of architectures and settings, but also largely closes the performance gap between them (\tblref{tab: lvis coco gap}). {\footnotesize{\demphinline{({timestamp: Nov 10, 2022})}}}}
\label{tab: lvis all}
\vspace{-8pt}
\end{table*}

\subsubsection{Object Detection \& Instance Segmentation}
\label{sec: object detection}

\paragraph{Datasets.} 
We evaluate the object detection and instance segmentation performance of \eva on both COCO~\cite{lin2014coco} and LVISv1.0~\cite{gupta2019lvis} benchmark.
COCO is a widely used object-level recognition benchmark consisting of 118k $\mathtt{train}$, 5k $\mathtt{val}$, and 20k $\mathtt{test}$-$\mathtt{dev}$ images respectively, with 80 common object categories.
LVISv1.0 is an emerging large-vocabulary object-level recognition benchmark, which has more than 1,200 object categories as well as more than 2 million high quality instance segmentation masks {\footnotesize{\demphinline{(nearly 2$\times$ of COCO instance masks)}}}.

Notably, COCO and LVISv1.0 almost use the same set of images, and both $\mathtt{train}$ and $\mathtt{val}$ split of LVISv1.0 have a huge overlap with COCO $\mathtt{train}$ and $\mathtt{val}$ split.
Meanwhile, COCO has much fewer object categories than LVISv1.0 (\ie, 80 \textit{v.s.}1,200+).
Therefore it is interesting and meaningful to evaluate a model's performance on both COCO and LVIS.

\paragraph{Metrics.} 
For COCO, we report the standard box AP (\boxAP) and mask AP (\maskAP) on both $\mathtt{val}$ and $\mathtt{test}$-$\mathtt{dev}$ split.
For LVISv1.0, we evaluate \eva using \boxAP, \maskAP and \maskAPrare defined in~\cite{gupta2019lvis} on the v1.0 $\mathtt{val}$ set.
We also report the experimental ``boundary, fixed AP''(\demphs{\boundaryAPfix} in \tblref{tab: lvis}) used in the LVIS 2021 challenge\footnote{Source: \url{https://www.lvisdataset.org/challenge_2021}} for reference.

\paragraph{Training \& evaluation settings.}
\eva uses Cascade Mask R-CNN~\cite{cai2019cascade} as the detector and adopts the training settings (\eg, large scale jitter data augmentation~\cite{simple_copy_paste}) \& architecture configurations (\eg, interleaved window \& global attention) of ViTDet~\cite{li2022exploring}.
Following the common practice~\cite{swinv2,detdino,beit3}, we first conduct intermediate fine-tuning for the whole detector using Objects365~\cite{o365} dataset with a resolution of 1024\suptext{2}, then we fine-tune the detector on COCO and LVISv1.0 $\mathtt{train}$ split respectively with 1280\suptext{2} inputs.

We report single-scale evaluation and multi-scale evaluation / test-time augmentation (tta) results of \eva for comparison.
For COCO, Soft-NMS~\cite{bodla2017soft} is also applied.
For instance segmentation task, the classification score is calibrated~\cite{huang2019mask} via maskness~\cite{solo}.

The model architecture as well as the hyper-parameters for COCO and LVISv1.0 are almost the \textit{same} {\footnotesize{\demphinline{(\ie, the hyper-parameters are nearly ``zero-shot'' transferred from COCO to LVISv1.0)}}}, expect we use federated loss~\cite{zhou2021probabilistic} and repeat factor sampling~\cite{gupta2019lvis} following ViTDet on LVISv1.0.

\paragraph{COCO Results.}
Perhaps COCO is the most fierce vision benchmark.
\tblref{tab: who he sota} compares \eva with some previous leading approaches on COCO.
Our model creates new state-of-the-art results on both object detection and instance segmentation tasks.

Compared with ViTDet-H~\cite{li2022exploring} that uses Cascade Mask R-CNN~\cite{cai2019cascade} as well, \eva shows that with a larger model and better encoder \& detector pre-training, the performance can be greatly improved with the same detector.

Compared with FocalNet~\cite{yang2022focal} and Group DETRv2~\cite{chen2022group} that choose better-established and highly-optimized DINO detector~\cite{detdino}, \eva demonstrates that with sufficient model size, data and pre-training, better performance can be also achieved via the classic R-CNN framework~\cite{rcnn}.
On the other hand, FocalNet and Group DETRv2 are incapable of instance segmentation due to using DINO.

Compared with SwinV2-Giant~\cite{swinv2} and FD-SwinV2-Giant~\cite{wei2022featdistill} that also adopt a {\footnotesize{\demphinline{(stronger HTC++~\cite{chen2019htc})}}} detector from the R-CNN family but with \app3$\times$ model size of \eva, our approach streamlines the pre-training processes and pulls off a ``Giant-killing'' act via better representations.

Compared with BEiT-3, \eva shows that is possible to build a state-of-the-art object-level recognition system without exploiting {\textbf{(i)}} semantic feature quantization / tokenization~\cite{bao2021beit,beitv2}, and {\textbf{(ii)}} image-text paired pre-training data and large corpora during pre-training.

\paragraph{LVIS Results.}
\tblref{tab: lvis} summarizes the results on LVISv1.0 $\mathtt{val}$ set. 
\eva achieves state-of-the-art performance under all metrics with single-scale evaluation, outperforming the previous best approaches by a large margin.

\paragraph{Analysis the LVIS-COCO gap: more is different.} 
It is interesting and valuable to evaluate and analyze a model on \textit{both} COCO and LVISv1.0 benchmarks, as they share almost the same image set but with different numbers of annotated object categories.
Compared with COCO which has only 80 annotated categories, LVIS annotates more than 1,200 object categories and thus naturally features a long-tail distribution, which is more close to the challenging real world scenario~\cite{gupta2019lvis}.
In general, LVIS is considered to be a much harder benchmark than COCO in object-level recognition, and conventional methods usually suffer from a large performance drop on LVIS compared with COCO.

In \tblref{tab: lvis coco gap}, we analyze the performance gap between LVISv1.0 and COCO benchmark of \eva and other state-of-the-art approaches. 
For previous leading methods such as ViTDet, the performance gap between \boxAP is around 8, and the gap between \maskAP is around 5.
However, using the same detector {\footnotesize{\demphinline{(Cascade Mask R-CNN)}}} and almost the same settings as ViTDet pre-trained via MAE-Huge {\footnotesize{\demphinline{(ViTDet-H)}}}, \eva not only achieves the state-of-the-art results on both LVIS \& COCO benchmarks simultaneously, but also largely closes the performance gap between them, especially for the instance segmentation task, that \eva achieves the \textit{same} performance on LVIS and COCO with single-scale evaluation.
Compared with ViTDet-H, we show a little larger model and stronger representations can take a great leap in the challenging large vocabulary instance segmentation benchmark--with one caveat described next.

Notice that it is inaccurate to say \eva ``solves'' the LVIS large vocabulary instance segmentation task based on ``zero \maskAP gap'', and it is possible to achieve even \textit{higher} \maskAP on LVIS than COCO {\footnotesize{\demphinline{(\eg, LVIS has higher quality mask annotations for both training and evaluation)}}}.
Although the \maskAPrare of \eva on LVIS is much better than previous approaches, there is still a big gap between the rare categories and common  / frequent categories {\footnotesize{\demphinline{(\ie, \maskAPrare = 48.3 \textit{v.s.} \maskAPcomm = 55.5 / \maskAPfreq = 57.4)}}}.
From another perspective, the gap of \boxAP between LVIS and COCO {\footnotesize{\demphinline{(1.9 for \eva)}}} may better reflect the overall gap between the large vocabulary and common vocabulary object-level recognition, since this metric eliminates the confounder of instance mask annotation quality.

Nevertheless, \eva makes a significant breakthrough in the challenging large vocabulary object-level recognition task.
We believe other large vision foundation models such as SwinV2-G and BEiT-3 also have similar properties on COCO and LVISv1.0 benchmark.
We hope our efforts can encourage the community to pay more attention to the relationships between different tasks while scaling up models and chasing the state-of-the-art in each individual task.

%#################################################
% semantic segmentation
%#################################################
\begin{table}[!t] 
% \vspace{-10pt}
    \centering
    \tablestyle{5.5pt}{1.15}
    \begin{tabular}{l|c|cc|c}
        & & \multicolumn{2}{c|}{\scriptsize{ADE20K}} & \scriptsize{COCO-Stuff} \\
        model & crop size & \mIoUss & \mIoUms & \mIoUss \\
        \shline
        HorNet~\cite{rao2022hornet} & 640\suptext{2} & 57.5 & 57.9 & - \\
        SeMask~\cite{jain2021semask} & 640\suptext{2} & 57.0 & 58.3 & - \\
        SwinV2-G~\cite{swinv2} & 896\suptext{2} & 59.3 & 59.9 & - \\
        Mask DINO~\cite{maskdino} & 896\suptext{2} & 59.5 & 60.8 & - \\
        FD-SwinV2-G~\cite{wei2022featdistill} & 896\suptext{2} & - & 61.4 & - \\
        ViT-Adapter~\cite{vitadapt} & 896\suptext{2} & 61.2 & 61.5 & 52.3 \\
        BEiT-3~\cite{beit3} & 896\suptext{2} & 62.0 & 62.8 & - \\ 
        \bf\evapurple{\eva} & 896\suptext{2} & \bf\evagreen{61.5} & \bf\evagreen{62.3} & \bf\evagreen{53.4} \\
        \end{tabular}
        \vspace{-5pt}
        \caption{\textbf{Semantic segmentation performance on ADE20K and COCO-Stuff-164K dataset.} ``\mIoUss'': mIoU of single-scale evaluation, ``\mIoUms'': mIoU using multi-scale evaluation.}
        % \vspace{-15pt}
        \label{tab: sem seg}
\end{table}

\subsubsection{Semantic Segmentation}
\label{sec: semantic segmentation}

\paragraph{Dataset.}
We evaluate \eva on ADE20K~\cite{zhou2018ade} and COCO-Stuff-164K~\cite{coco-stuff} datasets for semantic segmentation task.
ADE20K includes 150 semantic categories, and has 20k images for training \& 2k images for validation.
COCO-Stuff-164K augments 164K complex images from COCO with pixel-level annotations that span over 172 categories including 80 things, 91 stuff, and 1 unlabeled class.
Compared with ADE20K, COCO-Stuff is a more challenging but under-explored semantic segmentation benchmark.

\paragraph{Training \& evaluation settings.}
We follow the task transfer pipelines of ViT-Adapter~\cite{vitadapt}+mask2former~\cite{mask2former} but with a weakened model adaptation processes due to GPU memory limitation {\footnotesize{\demphinline{(40GB of VRAM)}}}:
\textbf{(i)} relative position biases~\cite{shaw2018self} are not applied. 
\textbf{(ii)} We use 8$\times$ decoders in mask2former segmentation head instead of 9\x. 
\textbf{(iii)} The feature dimension in mask2former head is \app0.6$\times$ of \eva encoder.

\paragraph{Results.}
We compare \eva with other leading semantic segmentation methods in \tblref{tab: sem seg}.
\eva achieves strong results in both ADE20K and COCO-Stuff-164K datasets.
On the other hand, the segmentation performance of \eva is slightly lower compared with BEiT-3 on ADE20K, we suspect this is partially due to our weakened architectural configurations.

%#################################################
% clip
%#################################################
\begin{table*}[t!]
% \vspace{-10pt}
\centering
%#################################################
% cfg
%#################################################
\subfloat[
\textbf{CLIP model configurations.} \eva CLIP-g can be stably trained via $\mathtt{fp16}$ precision with fewer image-text pairs (7B \textit{v.s.} 12B / 32B) sampled from a smaller data pool (LAION-400M \textit{v.s.} LAION-2B) on \app1/3\x GPUs compared with other open-sourced billion-scale competitors.
\label{tab: clip cfg}
]{
\centering
\begin{minipage}{1\linewidth}{\begin{center}
    \tablestyle{3.5pt}{1.2}
    \begin{tabular}{r|c|ccc|cc|cccl}
        & & total & image & text & & & & & \\
        model & precision & \#param. & \#param. & \#param. & clip training data & samples seen & image size & patch size & batch size & gpus for training \\
        \shline
        \demphs{OpenAI CLIP-L} & \demphs{$\mathtt{float16}$} & \demphs{430M} & \demphs{304M} & \demphs{124M} & \scriptsize{\demphs{CLIP-400M}~\cite{clip}} & \demphs{12B} & \demphs{224\suptext{2}} & \demphs{14{\scriptsize{\x}}14} & \demphs{32k} & \demphs{256{\scriptsize{\x}}V100 \scriptsize{(32GB)}} \\
        \demphs{ALIGN} & \demphs{$\mathtt{bfloat16}$} & \demphs{834M} & \demphs{480M} & \demphs{354M} & \scriptsize{\demphs{ALIGN-1.8B}~\cite{clip}} & \demphs{22B} & \demphs{289\suptext{2}} & \demphs{-} & \demphs{16k} & \demphs{1024{\scriptsize{\x}}TPUv3} \\
        Open CLIP-H & $\mathtt{bfloat16}$ & 1.0B & 632M & 354M & \scriptsize{LAION-2B~\cite{laion5b}} & 32B & 224\suptext{2} & 14{\scriptsize{\x}}14 & 79k & 824{\scriptsize{\x}}A100 \scriptsize{(40GB)} \\
        Open CLIP-g & $\mathtt{bfloat16}$ & 1.3B & 1.0B & 354M & \scriptsize{LAION-2B~\cite{laion5b}} & 12B & 224\suptext{2} & 14{\scriptsize{\x}}14 & 64k & 800{\scriptsize{\x}}A100 \scriptsize{(40GB)} \\
        \textbf{\evapurple{\eva}} CLIP-g & $\mathtt{float16}$ & 1.1B & 1.0B & 124M & \scriptsize{LAION-400M~\cite{laion400m}} & 11B & 224\suptext{2} & 14{\scriptsize{\x}}14 & 41k & 256{\scriptsize{\x}}A100 \scriptsize{(40GB)} \\
        \end{tabular}
\end{center}}\end{minipage}
}
\vspace{1.em}
\\
%#################################################
% clip result
%#################################################
\subfloat[
\textbf{Summary of zero-shot image / video classification performance.} ``{$\Delta$\scriptsize{$\downarrow$}}'': The gap between the averaged performance of ImageNet-\{1K, V2, Adv., Ren., Ske.\} \& ObjectNet that with natural distribution shifts and the original ImageNet-1K validation accuracy. Our model suffers from the smallest performance drop (only \textbf{\evagreen{2.5}}\% top-1 accuracy gap) while maintaining the highest zero-shot classification accuracy averaged on all 12 benchmarks (\textbf{\evagreen{72.7}}\% top-1 accuracy).
\label{tab: clip result}
]{
\centering
\begin{minipage}{1\linewidth}{\begin{center}
\tablestyle{2.5pt}{1.2}
    \begin{tabular}{r|llllllll|c|llll|l}
        \rotatebox[origin=l]{90}{datasets} &
        \ph{+}\rotatebox[origin=l]{90}{\scriptsize{ImageNet-1K~\cite{deng2009imagenet}}} &
        \ph{+}\rotatebox[origin=l]{90}{\scriptsize{ImageNet-V2~\cite{recht2019imagenetv2}}} &
        \ph{+}\rotatebox[origin=l]{90}{\scriptsize{ImageNet-Adv.~\cite{inadv}}} &
        \ph{+}\rotatebox[origin=l]{90}{\scriptsize{ImageNet-Ren.~\cite{inren}}} &
        \ph{+}\rotatebox[origin=l]{90}{\scriptsize{ImageNet-Ske.~\cite{inske}}} &
        \ph{+}\rotatebox[origin=l]{90}{\scriptsize{ObjectNet~\cite{objectnet}}} &
        \ph{+}\rotatebox[origin=l]{90}{\scriptsize{CIFAR-10~\cite{cifar}}} &
        \ph{+}\rotatebox[origin=l]{90}{\scriptsize{CIFAR-100~\cite{cifar}}} & 
        &
        \ph{+}\rotatebox[origin=l]{90}{\scriptsize{UCF-101~\cite{ucf101}}} &
        \ph{+}\rotatebox[origin=l]{90}{\scriptsize{Kinetics-400~\cite{kay2017kinetics}}} & 
        \ph{+}\rotatebox[origin=l]{90}{\scriptsize{Kinetics-600~\cite{k600}}} & 
        \ph{+}\rotatebox[origin=l]{90}{\scriptsize{Kinetics-700~\cite{k700}}} &
        \\
        \shline
        model & \multicolumn{8}{c|}{\scriptsize{image classification}} & $\Delta$\scriptsize{$\downarrow$} & \multicolumn{4}{c|}{\scriptsize{video classification}} & avg. all \\
        \hline
        \demphs{OpenAI CLIP-L} 
        & \scriptsize{\demphs{75.5}} 
        & \scriptsize{\demphs{69.9}} 
        & \scriptsize{\demphs{70.8}} 
        & \scriptsize{\demphs{87.8}} 
        & \scriptsize{\demphs{59.6}} 
        & \scriptsize{\demphs{69.0}} 
        & \scriptsize{\demphs{95.6}} 
        & \scriptsize{\demphs{75.9}} 
        & \scriptsize{3.4} 
        & \scriptsize{\demphs{76.4}} 
        & \scriptsize{\demphs{64.5}} 
        & \scriptsize{\demphs{64.2}} 
        & \scriptsize{\demphs{57.7}} 
        & \scriptsize{\demphs{72.2}} \\
        \demphs{ALIGN} 
        & \scriptsize{\demphs{76.4}} 
        & \scriptsize{\demphs{70.1}} 
        & \scriptsize{\demphs{75.8}} 
        & \scriptsize{\demphs{92.2}} 
        & \scriptsize{\demphs{64.8}} 
        & \scriptsize{\demphs{72.2}} 
        & \scriptsize{\demphs{-}} 
        & \scriptsize{\demphs{-}} 
        & \scriptsize{-} 
        & \scriptsize{\demphs{-}} 
        & \scriptsize{\demphs{-}} 
        & \scriptsize{\demphs{-}} 
        & \scriptsize{\demphs{-}} 
        & \scriptsize{\demphs{-}} \\
        Open CLIP-H 
        & \scriptsize{78.0} 
        & \scriptsize{70.9} 
        & \scriptsize{59.3} 
        & \scriptsize{89.3} 
        & \scriptsize{66.6} 
        & \scriptsize{69.7} 
        & \scriptsize{97.5} 
        & \scriptsize{84.7} 
        & \scriptsize{5.7} 
        & \scriptsize{78.2} 
        & \scriptsize{63.1} 
        & \scriptsize{63.6} 
        & \scriptsize{56.1} 
        & \scriptsize{73.1} \\
        Open CLIP-g 
        & \scriptsize{76.6} 
        & \scriptsize{69.6} 
        & \scriptsize{57.2} 
        & \scriptsize{88.7} 
        & \scriptsize{65.2} 
        & \scriptsize{67.5} 
        & \scriptsize{97.1} 
        & \scriptsize{83.9} 
        & \scriptsize{5.8} 
        & \scriptsize{77.7} 
        & \scriptsize{61.7} 
        & \scriptsize{62.2} 
        & \scriptsize{55.0} 
        & \scriptsize{71.9} \\
        \textbf{\evapurple{\eva}} CLIP-g 
        & \scriptsize{\textbf{78.5}\dtplus{+1.9}} 
        & \scriptsize{\textbf{71.5}\dtplus{+1.9}} 
        & \scriptsize{\textbf{73.6}\dtplus{+16.4}} 
        & \scriptsize{\textbf{92.5}\dtplus{+3.8}} 
        & \scriptsize{\textbf{67.3}\dtplus{+2.1}} 
        & \scriptsize{\textbf{72.3}\dtplus{+4.8}} 
        & \scriptsize{\textbf{98.3}\dtplus{+1.2}} 
        & \scriptsize{\textbf{88.7}\dtplus{+4.8}} 
        & \scriptsize{\textbf{\evagreen{2.5}}} 
        & \scriptsize{\textbf{76.1}\dt{-1.6}} 
        & \scriptsize{\textbf{65.2}\dtplus{+3.5}} 
        & \scriptsize{\textbf{64.4}\dtplus{+2.2}} 
        & \scriptsize{\textbf{58.4}\dtplus{+3.4}} 
        & \scriptsize{\textbf{75.7}\dtplus{+3.8}}
        \end{tabular}
\end{center}}\end{minipage}
}
\vspace{-.3em}
\caption{\textbf{\eva as a vision-centric, multi-modal pivot.} We evaluate a billion-scale contrastive language-image pre-trained (CLIP) model with the vision tower initialized from pre-trained \eva, which largely accelerates the contrastive training efficiency and shows promising zero-shot classification performance across a wide range of image / video benchmarks. The statistics \& performance of \eva's MIM teacher {\footnotesize{\demphinline{({OpenAI CLIP-L})}}} are also presented for reference. A thorough evaluation of \textbf{\evapurple{\eva}} CLIP on 35 zero-shot benchmarks {\footnotesize{\demphinline{(including 27 zero-shot image classification benchmarks + 4 zero-shot video classification benchmarks + 2\x2 zero-shot retrieval benchmarks)}}} is available at \href{https://github.com/baaivision/EVA/blob/master/clip/README.md}{link1} and \href{https://github.com/baaivision/EVA/blob/master/clip/benchmark.md}{link2}.}
\label{tab: clip}
\vspace{-10pt}
\end{table*}

\subsubsection{Contrastive Language-Image Pre-training \\ with Zero-shot Classification Evaluation}
\label{sec: clip exp}

CLIP (Contrastive Language-Image Pre-training)~\cite{clip,align,basic,openclip} is a type of multi-modal foundation model that connects vision and language via contrastive image-text pre-training.
CLIP can be applied to any image classification benchmark by simply providing the names of the visual categories to be recognized~\cite{clipblog}.
Thus the introduction of CLIP essentially reshapes the landscape of visual recognition.
Meanwhile, CLIP features also play a central role in representation leaning~\cite{beitv2,beit3}, AI generated content~\cite{dalle2,imagen,stablediffusion} and large dataset filtering~\cite{kakaobrain2022coyo-700m,laion400m,laion5b}, \etc.

In this section and \tblref{tab: clip}, we show that \eva is not only a strong encoder for a wide range of vision downstream tasks, but also a multi-modal pivot that builds a bridge between vision and language.
To demonstrate that, we train \& evaluate \eva as a billion-scale CLIP's vision tower in various zero-shot image / video classification benchmarks.

\paragraph{Baselines and major challenges in CLIP model scaling.}
We compare our CLIP (dubbed \eva CLIP) with other open-sourced strong CLIP competitors that exploit publicly accessible data / academic resources only.
Model configurations and statistics are detailed in \tblref{tab: clip cfg}.

There are two well-known major challenges of CLIP model training and scaling: 
\textbf{(i)} Large-scale Open CLIP models (\eg, Open CLIP-H \& Open CLIP-g~\cite{openclip,laiongiantclip}) usually suffer from severe training instability issues~\cite{laiongiantclip} and have to use $\mathtt{bfloat16}$ format for optimization.
\textbf{(ii)} The training efficiency is low, which may hinder model scaling and downstream performance.
For instance, Open CLIP-g is heavily under-trained due to its large compute requirement, and its performance is \textit{even worse} than the sufficiently-trained Open CLIP-H with a smaller model size.

Compared with our CLIP model, Open CLIP-H \& -g are trained from scratch with much more image-text pairs {\footnotesize{\demphinline{(\app2.9$\times$ and \app1.1$\times$ of ours)}}} sampled from a much larger dataset {\footnotesize{\demphinline{(\app5$\times$ of ours)}}} on \app3$\times$ of GPUs.
While by leveraging \eva, billion-scale CLIP model training can be accelerated with improved zero-shot classification performance, described next.

\paragraph{Training settings.}
For our CLIP model, we initialize the vision encoder via pre-trained \eva and the language encoder from OpenAI CLIP-L.
The pre-training implementation is based on Open CLIP~\cite{openclip}.
We also adopt $\mathtt{DeepSpeed}$ optimization library~\cite{rasley2020deepspeed} with ZeRO stage-1 optimizer~\cite{rajbhandari2020zero} to save memory.
We find using $\mathtt{fp16}$ format with dynamic loss scaling is stable enough during the whole course of training while using $\mathtt{bfloat16}$ format is unnecessary.
These modifications allow us to train a 1.1B CLIP with a batch size of 41k on 256$\times$ NVIDIA A100 40GB GPUs.

\paragraph{Evaluation settings.}
We evaluate zero-shot image / video classification performance of each CLIP model on 12 benchmarks and report top-1 accuracy for comparisons.

For zero-shot image classification task, we choose 8 benchmarks, \ie, ImageNet-1K~\cite{deng2009imagenet}, ImageNet-V2~\cite{inv2}, ImageNet-Adversarial {\footnotesize{\demphinline{(ImageNet-Adv.)}}}~\cite{inadv}, ImageNet-Rendition {\footnotesize{\demphinline{(ImageNet-Adv.)}}}~\cite{inren}, ImageNet-Sketch {\footnotesize{\demphinline{(ImageNet-Ske.)}}}~\cite{inske}, ObjectNet~\cite{objectnet}, CIFAR-10 and CIFAR-100~\cite{cifar}.
We are also interested in the robustness of CLIP models, evaluated via the performance gap between the averaged performance of ImageNet-\{1K, V2, Adv., Ren., Ske.\} \& ObjectNet that with natural distribution shifts and the original ImageNet-1K validation accuracy.

For zero-shot video classification task, we choose 4 benchmarks, namely UCF-101~\cite{ucf101}, Kinetics-400~\cite{kay2017kinetics}, Kinetics-600~\cite{k600}, and Kinetics-700~\cite{k700}.

\paragraph{Results.}
\tblref{tab: clip result} shows the comparison.
Our \eva CLIP achieves the highest averaged accuracy, and performs the best in 10 out of 12 zero-shot classification benchmarks.
Notably, the ImageNet-1K validation zero-shot top-1 accuracy is 78.2\% without
using any of its training set labels, matching the original ResNet-101~\cite{resnet}.
Moreover, our model is quite robust and suffers from the smallest performance drop when facing natural distribution shifts in ImageNet.

At last, in \tblref{tab: eva-clip 1k cls} we provide zero-shot, linear probing \& end-to-end fine-tuning top-1 accuracy of \eva-CLIP on ImageNet-1K validation set for reference.
Our approach creates the new state-of-the-art results among all existing self-supervised learning methods.

\begin{table}[h!]
\centering
\tablestyle{6pt}{1.15}
% \scriptsize
\begin{tabular}{y{40}|x{50}x{50}x{50}}
model \scriptsize{(SSL)} & zero-shot & linear probing & fine-tuning \\
\shline
prev. best & 78.0\suptext{\demphs{a}} & 82.3\suptext{\demphs{b}} & 89.1\suptext{\demphs{c}} \\
\eva & \bf\evagreen{78.5}\ph{\suptext{a}} & \bf\evagreen{86.5}\ph{\suptext{b}} & \bf\evagreen{89.4}\ph{\suptext{c}} \\
\end{tabular}
\caption{\textbf{Zero-shot, linear probing and fine-tuning} performance of \eva-CLIP on ImageNet-1K. Notice that the linear probing and fine-tuning results are from the vision encoder of \eva-CLIP. Our approach establishes the new state-of-the-art results among all existing self-supervised learning (SSL) methods. {\footnotesize{\demphinline{({timestamp: Nov 10, 2022})}}} {\scriptsize results reference. a: Open CLIP-H~\cite{openclip}, b: iBOT~\cite{zhou2021ibot}, c: dBOT~\cite{liu2022exploring}.}}
\label{tab: eva-clip 1k cls}
\end{table}

Notice that \eva CLIP's vision branch learns from OpenAI CLIP-L, while language branch initialized from the same CLIP-L model.
Therefore, starting from a CLIP-L with only 430M parameters, we progressively scale up a 1.1B \eva CLIP-g with large performance improvements.
This implies that interleaved MIM \& image-text contrastive pre-training could be an efficient and scalable CLIP training approach.
To our knowledge, \eva CLIP-g is the largest performant CLIP model trained via publicly accessible data and resources.
We hope our practice on scaling and improving CLIP can also inspire and transfer to the study of other large scale multi-modal foundation models.

\section{Related Work}
\label{sec:relatedwork}

\paragraph{Masked image modeling (MIM)} learns rich visual representations via predicting masked visual contents conditioned on visible context.
ViT~\cite{dosovitskiy2020vit} and iGPT~\cite{igpt} report the first meaningful MIM pre-training results.
The BEiT family~\cite{bao2021beit, beitv2, beit3} greatly improves MIM's performance via masked visual token prediction.
Recent work~\cite{he2021masked, xie2021simmim, wei2022masked, zhou2021ibot, chen2022context, fang2022corrupted, baevski2022data2vec, dong2021peco} (re-)explore pixel / feature regression in MIM, but only in a relatively small model and data scales.
In this work, we explore the limits of large scale MIM pre-training via masked image-text aligned feature prediction~\cite{wei2022mvp, hou2022milan}.

\paragraph{Vision foundation models.} 
ConvNets~\cite{cnn} have long been the de-facto standard visual architecture ab initio.
Since AlexNet~\cite{alexnet}, ConvNets have rapidly evolved and become deeper, wider and larger~\cite{vgg, inception, resnet, resnext, densenet, efficientnet, convnext}.
However, at sufficient model and data scales, ConvNets lag behind ViTs~\cite{dosovitskiy2020vit} due to a lack of scalable pre-training tasks and the built-in inductive biases.
Entering the 2020s, large pre-trained ViTs~\cite{dosovitskiy2020vit, zhai2022scalingvit} such as SwinV2-G~\cite{swinv2} with hierarchical architectures as well as BEiT-3~\cite{beit3} with multi-modal representations started to demonstrate various vision benchmarks.
In this work, we show by leveraging unlabeled images, vanilla ViT can be efficiently scaled up to billion-scale parameters, and stands out in various downstream tasks.

\section{Conclusion}
\label{sec:conclusion}
In this work, we launch \eva, a one billion parameters vanilla ViT encoder to explore the limits of masked visual representation learning.
We show simple masked feature modeling as a visual learning pretext task scales well on an architecture with minimal vision priors, and attains excellent
results in a representative \& diverse set of downstream tasks.
We hope \eva would bridge the gap between vision and language study via masked modeling, and contributes to the \evapurple{\textbf{Neon Genesis}} of vision research.

\section*{Acknowledgement}
\label{sec:ack}
We would like to thank Hanxiao Qu, Yan Tian, Yemin Shi and Xigang Cao for their help on GPU resources.
Zhao Xue, Quanyue Ma and Bowen Zhang for their help on datasets and benchmarks, and other colleagues at Beijing Academy of Artificial Intelligence for support throughout this project.

\appendix

\section{Appendix}
\label{app}

The MIM pre-training and contrastive language-image pre-training settings are already available in our main submission.
Here we summarize the detailed configurations for image classification (\sref{app: img cls}), video action classification (\sref{app: vid cls}), object detection \& instance segmentation (\sref{app: det}), and semantic segmentation (\sref{app: sem seg}).

\subsection{Image Classification}
\label{app: img cls}

The fine-tuning hyper-parameters for ImageNet-21K and ImageNet-1K are shown in \tblref{app tab: 21k cls} and \tblref{app tab: 1k cls}, respectively.

%##################################################################################################
\begin{table}[h]
% \vspace{-.5em}
\centering
\tablestyle{6pt}{1.15}
\scriptsize
\begin{tabular}{y{96}|x{96}}
config & value \\
\shline
peak learning rate & 1e-4 \\
optimizer & AdamW~\cite{adam,Loshchilov2019adamw} \\
optimizer hyper-parameters & $\beta_1$, $\beta_2$, $\epsilon$ = 0.9, 0.98, 1e-6 \\
layer-wise lr decay~\cite{clark2020electra, bao2021beit} & 0.85 \\
learning rate schedule & cosine decay \\
weight decay & 0.05 \\
\hline
input resolution & 224\suptext{2} \\
batch size & 4096 \\
warmup epochs & 15 \\
training epochs & 60 \\
\hline
drop path~\cite{huang2016deep} & 0.4 \\
augmentation & RandAug (9, 0.5)~\cite{cubuk2020randaugment} \\
label smoothing~\cite{szegedy2016rethinking} & 0.1 \\
cutmix~\cite{yun2019cutmix} & 1.0 \\
mixup~\cite{zhang2017mixup} & \xmark \\
random erasing~\cite{zhong2020random} & \xmark \\
random resized crop & (0.5, 1) \\
ema & \xmark
\end{tabular}
\vspace{-.5em}
\caption{Intermediate fine-tuning setting for ImageNet-21K.}
% \vspace{-.5em}
\label{app tab: 21k cls}
\end{table}
%##################################################################################################

%##################################################################################################
\begin{table}[!t]
% \vspace{-15pt}
\centering
\tablestyle{6pt}{1.15}
\scriptsize
\begin{tabular}{y{96}|x{96}}
config & value \\
\shline
peak learning rate & 3e-5 \\
optimizer & AdamW \\
optimizer hyper-parameters & $\beta_1$, $\beta_2$, $\epsilon$ = 0.9, 0.999, 1e-8 \\
layer-wise lr decay & 0.95 \\
learning rate schedule & cosine decay \\
weight decay & 0.05 \\
\hline
input resolution & 336\suptext{2} / \evapurple{560\suptext{2}} \\
batch size & 512 \\
warmup epochs & 2 \\
training epochs & 10 / \evapurple{15} \\
\hline
drop path & 0.4 \\
augmentation & RandAug (9, 0.5) \\
label smoothing & 0.3 \\
cutmix & \xmark \\
mixup & \xmark \\
random erasing & \xmark \\
random resized crop & (0.08, 1) \\
ema & 0.9999 \\
test crop ratio & 1.0
\end{tabular}
\vspace{-.5em}
\caption{Fine-tuning setting for ImageNet-1K.}
% \vspace{-.5em}
\label{app tab: 1k cls}
\end{table}
%##################################################################################################

\subsection{Video Action Classification}
\label{app: vid cls}

%#################################################
% video data statics
%#################################################
\begin{table}[!b] 
    \centering
    \tablestyle{6pt}{1.2}
    \scriptsize
    \begin{tabular}{y{70}|z{30}x{50}x{30}}
        dataset \& split & \#clips & avg. length & \#classes \\
        \shline
        Kinetics-400 train~\cite{kay2017kinetics} & 234,584 & 10s & 400 \\
        Kinetics-400 val~\cite{kay2017kinetics} & 19,760 & 10s & 400 \\
        % \hline
        Kinetics-600 train~\cite{k600} & 412,688 & 10s & 600 \\
        Kinetics-600 val~\cite{k600} & 29,779 & 10s & 600 \\
        % \hline
        Kinetics-700 train~\cite{k700} & 534,063 & 10s & 700 \\
        Kinetics-700 val~\cite{k700} & 33,914 & 10s & 700 \\
        \hline
        Kinetics-722 (ours) & 629,395 & 10s & 722 \\
        \end{tabular}
        \vspace{-.5em}
        \caption{Video dataset statistics.}
        \label{app tab: video stat}
\end{table}

%#################################################
% video fine-tune settings
%#################################################
\begin{table}[h!]
% \vspace{-15pt}
\centering
\tablestyle{6pt}{1.15}
\scriptsize
\begin{tabular}{y{96}|x{96}}
config & value \\
\shline
optimizer & AdamW \\
optimizer hyper-parameters & $\beta_1$, $\beta_2$, $\epsilon$ = 0.9, 0.98, 1e-6 \\
weight decay & {0.05} \\
peak learning rate & 8e-6 \\
learning rate schedule & cosine decay \\
warmup epochs & {5} \\
epochs & {40} \\
batch size & 256 \\
\hline
input resolution & 224\suptext{2} \\
random flip & 0.5 \\ 
multiscale crop & (1, 0.875, 0.75, 0.66) \\
color jitter & 0.8 \\ 
grayscale & 0.2 \\ 
cutmix & 1.0 \\
mixup & 0.8 \\
label smoothing & 0.1 \\
drop path & 0.3 \\
layer-wise lr decay & \xmark \\
\end{tabular}
\vspace{-.5em}
\caption{Kinetics-722 intermediate fine-tuning settings.}
\label{app tab: k722 param}
\end{table}

%#################################################
% video fine-tune settings
%#################################################
\begin{table}[t!]\centering
\tablestyle{3pt}{1.15}
\scriptsize
\begin{tabular}{y{85}|x{35}x{35}x{35}}
config & {K-400~\cite{kay2017kinetics}} & {K-600~\cite{k600}} & {K-700~\cite{k700}} \\
\shline
optimizer & \multicolumn{3}{c}{AdamW} \\
optimizer hyper-parameters & \multicolumn{3}{c}{$\beta_1$, $\beta_2$, $\epsilon$ = 0.9, 0.98, 1e-6} \\
weight decay & \multicolumn{3}{c}{0.05} \\
peak learning rate & \multicolumn{3}{c}{1e-6} \\
minimal learning rate & \multicolumn{3}{c}{1e-6} \\
warmup epochs & \multicolumn{3}{c}{0} \\
epochs & 1 & 2 & 2 \\
batch size & \multicolumn{3}{c}{256}  \\
\hline
input resolution & \multicolumn{3}{c}{224\suptext{2}} \\
random flip & \multicolumn{3}{c}{0.5} \\ 
multiscale crop & \multicolumn{3}{c}{(1, 0.875, 0.75, 0.66)} \\
color jitter & \multicolumn{3}{c}{0.8} \\ 
grayscale & \multicolumn{3}{c}{0.2} \\ 
mixup & \multicolumn{3}{c}{\xmark} \\
cutmix & \multicolumn{3}{c}{\xmark} \\
label smoothing & \multicolumn{3}{c}{0.1} \\
drop path & \multicolumn{3}{c}{0.2} \\
layer-wise lr decay & \multicolumn{3}{c}{0.95}  \\
multi-view inference & \multicolumn{3}{c}{4 clips, 3 crops}
\end{tabular}
\vspace{-.5em}
\caption{Hyper-parameters used in the video action recognition. }
\label{app tab: k467 param}
\vspace{-10pt}
\end{table}

For video action classification tasks, a two-stage fine-tuning process is adopted.
The statistics of video datasets we used are available in \tblref{app tab: video stat}.

In the first stage, we conduct intermediate fine-tuning on a merged dataset coined Kinetics-722 (K-722) that integrates all valid training samples from Kinetics-400 (K-400)~\cite{kay2017kinetics}, Kinetics-600 (K-600)~\cite{k600} and Kinetics-700 (K-700)~\cite{k700}. 
The input video resolution is 224\suptext{2} with 8 frames.
Notably, for a fair and legal comparison, we removed leaked videos in all validation sets and duplicated videos in all training sets based on the videos' ``youtube id". 
Accordingly, the cleaned K-722 contains 0.63M training videos, covering 722 human action classes.
\tblref{app tab: k722 param} lists the detailed settings \& hyper-parameters for fine-tuning on this dataset.

In the second stage, we further fine-tune on each dataset using more input video frames of 16 with a resolution of 224\suptext{2}. For the frame sampling, we adopt the sparse sampling strategy~\cite{wang2016temporal}.
During testing, we follow the common practice of multi-view inference~\cite{tong2022videomae,liu2022video,feichtenhofer2022masked,wei2022masked} with 4 temporal clips and 3 spatial crops. 
The final prediction is the ensemble of all trials.
\tblref{app tab: k467 param} lists the detailed hyper-parameters for fine-tuning on K-400, K-600 and K-700.

\subsection{Object Detection \& Instance Segmentation}
\label{app: det}

The detailed hyper-parameters are shown in \tblref{app tab: o365} and \tblref{app tab: coco lvis}.
For intermediate fine-tuning on Objects365~\cite{o365}, the model is trained with a batch size of 128 for 380k iterations.
To accelerate the training process, we use a smaller input resolution of 1024\suptext{2} for the first 320k iteration. 
Afterward, the input resolution is lifted to 1280\suptext{2} for a better adaptation to the fine-tuning of COCO and LVIS.

For fine-tuning COCO and LVIS, the learning rate is initialized as 2.5e-5 and step by a factor of 10 for the last 5k iterations.
As shown in \tblref{app tab: coco lvis}, we use almost identical hyper-parameters for training COCO and LVIS. 
Except for the commonly used repeat factor sampling~\cite{gupta2019lvis} and federated loss~\cite{zhou2021probabilistic} that are specialized for long-tailed recognition, the only difference in training is that we train the model for 45k steps on COCO, while a longer 75k step on LVIS, since the tail classes generally take a longer schedule to converge~\cite{fu2021lvis}.

\begin{table}[h!]
\centering
\tablestyle{6pt}{1.15}
\scriptsize
\begin{tabular}{y{85}|x{100}}
config  & value \\
\shline
optimizer & AdamW \\
optimizer hyper-parameters & $\beta_1$, $\beta_2$, $\epsilon$ = 0.9, 0.999, 1e-8 \\
learning rate & 1e-4 \\
layer-wise lr decay & 0.9 \\
training steps & 380k \\
training input resolution & 1024\suptext{2} $\to$ 1280\suptext{2} \\
batch size & 128 \\
weight decay & {0.1} \\
% warmup steps & 15k \\
drop path & 0.6 \\
\end{tabular}
\caption{Objects365 object detection intermediate fine-tuning settings.}
\label{app tab: o365}
% \vspace{-10pt}
\end{table}

\begin{table}[h!]
\centering
\tablestyle{6pt}{1.15}
\scriptsize
\begin{tabular}{y{85}|x{43}x{43}}
config & {COCO} & {LVIS} \\
\shline
optimizer & \multicolumn{2}{c}{AdamW} \\
optimizer hyper-parameters & \multicolumn{2}{c}{$\beta_1$, $\beta_2$, $\epsilon$ = 0.9, 0.999, 1e-8} \\
learning rate & \multicolumn{2}{c}{2.5e-5} \\
learning rate schedule & \multicolumn{2}{c}{step decay} \\
training steps & {45k} & {75k} \\
learning decay step & 40k & 70k \\
batch size & \multicolumn{2}{c}{64}  \\
training input resolution & \multicolumn{2}{c}{1280\suptext{2}} \\
weight decay & \multicolumn{2}{c}{0.1} \\
layer-wise lr decay & \multicolumn{2}{c}{0.9} \\
% warmup steps & \multicolumn{2}{c}{250} \\
drop path & \multicolumn{2}{c}{0.6} \\
repeat threshold & - & 0.001 \\
frequency weight power & - & 0.5 \\
max numbers of detection & 100 & 1000 \\
\end{tabular}
\caption{COCO and LVIS object detection \& instance segmentation fine-tuning settings.}
\label{app tab: coco lvis}
% \vspace{-10pt}
\end{table}

\subsection{Semantic Segmentation}
\label{app: sem seg}

Detailed configurations about semantic segmentation are available in \tblref{app tab: sem seg}.
Our settings basically follow ViT-Adapter~\cite{vitadapt} with Mask2Former~\cite{mask2former} as the segmentation head.
For ADE20K, we use COCO-Stuff pre-trained weights as initialization.

\begin{table}[t!]
\centering
\tablestyle{6pt}{1.15}
\scriptsize
\begin{tabular}{y{85}|x{43}x{43}}
config& COCO-Stuff & ADE20K \\
\shline
optimizer & \multicolumn{2}{c}{AdamW} \\
optimizer hyper-parameters & \multicolumn{2}{c}{$\beta_1$, $\beta_2$, $\epsilon$ = 0.9, 0.999, 1e-8} \\
peak learning rate & 1.5e-5 & 2.5e-5 \\
batch size & 32 & 64 \\
fine-tuning steps & 60000 & 20000 \\
layer-wise lr decay & \multicolumn{2}{c}{0.95} \\
weight decay & \multicolumn{2}{c}{0.5} \\
drop path & \multicolumn{2}{c}{0.5} \\
input resolution & \multicolumn{2}{c}{896\suptext{2}} \\
seg head \#enc. \& \#dec. & \multicolumn{2}{c}{6 \& 8} \\
seg head dim & \multicolumn{2}{c}{1024} \\
relative position bias & \multicolumn{2}{c}{\xmark}
\end{tabular}
\caption{COCO-Stuff-164K and ADE20K semantic segmentation fine-tuning settings.}
\label{app tab: sem seg}
\end{table}

%%%%%%%%% REFERENCES
{
\fontsize{8.2pt}{9.84pt}\selectfont
\bibliographystyle{ieee_fullname}
\bibliography{egbib}
}

\end{document}